\PassOptionsToPackage{table}{xcolor}

\documentclass[acmsmall, screen, nonacm]{acmart}

\AtBeginDocument{%
  }

\acmDOI{10.1145/3742784}

\acmBooktitle{ACM Computing Surveys}
\usepackage{amsmath}
\usepackage{amsfonts}
\usepackage{amsthm}
\usepackage[table]{xcolor}
\usepackage{graphicx}
\usepackage{url}            
\usepackage{booktabs}       
\usepackage{amsfonts}       
\usepackage{nicefrac}       
\usepackage{microtype}      
\usepackage{xcolor}
\usepackage{tikz}
\usepackage{wrapfig}
\usepackage{enumitem}
\usepackage{multicol}
\usepackage{multirow}
\usepackage{wrapfig}
\usepackage{subcaption}
\usepackage{todonotes}
\usepackage{hyperref}

\theoremstyle{definition}
\newtheorem{example}{Example}%[section]
\definecolor{ForestGreen}{RGB}{34,139,34}

\usepackage{booktabs}
\newcommand{\bftab}[1]{\scalebox{.93}[1]{\bfseries #1}}

\usepackage[acronym, nohypertypes={acronym}]{glossaries}  

\newacronym{stgnn}{STGNN}{spatiotemporal graph neural network}
\newacronym{gnn}{GNN}{graph neural network}
\newacronym{mp}{MP}{message-passing}
\newacronym{mlp}{MLP}{multilayer perceptron}
\newacronym{tts}{TTS}{time-then-space}
\newacronym{stt}{STT}{space-then-time}
\newacronym{tas}{T\&S}{time-and-space}
\newacronym{rnn}{RNN}{recurrent neural network}
\newacronym{tcn}{TCN}{temporal convolutional network}
\newacronym{gru}{GRU}{gated recurrent unit}
\newacronym{sn}{SN}{\textit{sensor network}}
\newacronym{iot}{IoT}{Internet of Things}
\newacronym{stmp}{STMP}{spatiotemporal message-passing}
\newacronym{stcn}{STCN}{spatiotemporal convolutional network}
\newacronym{gcrnn}{GCRNN}{graph convolutional recurrent neural network}
\newacronym{mae}{MAE}{mean absolute error}
\newacronym{mape}{MAPE}{mean absolute percentage error}
\newacronym{rmse}{RMSE}{root mean squared error}
\newacronym{mre}{MRE}{mean relative error}

\glsdisablehyper
\newglossaryentry{ttsimp}{name=RNN\texttt{+}IMP,description=}
\newglossaryentry{ttsamp}{name=RNN\texttt{+}AMP,description=}
\newglossaryentry{tasimp}{name=GCRNN-IMP,description=}
\newglossaryentry{tasamp}{name=GCRNN-AMP,description=}

\newglossaryentry{fcrnn}{name=FC-RNN,description=}
\newglossaryentry{local_rnn}{name=LocalRNNs,description=}
\newglossaryentry{dcrnn}{name=DCRNN,description=}
\newglossaryentry{agcrn}{name=AGCRN,description=}
\newglossaryentry{gwnet}{name=GraphWaveNet,description=}

\newglossaryentry{gpvar}{name=GPVAR-G,description=}
\newglossaryentry{lgpvar}{name=GPVAR-L,description=}
\newglossaryentry{la}{name=METR-LA,description=}
\newglossaryentry{bay}{name=PEMS-BAY,description=}
\newglossaryentry{cer}{name=CER-E,description=}
\newglossaryentry{air}{name=AQI,description=}
\newglossaryentry{pems3}{name=PEMS03,description=}
\newglossaryentry{pems4}{name=PEMS04,description=}
\newglossaryentry{pems7}{name=PEMS07,description=}
\newglossaryentry{pems8}{name=PEMS08,description=}
\newglossaryentry{engrad}{name=EngRAD,description=}
\usepackage{amsmath, amsfonts, bm, tabstackengine}

\newcommand{\autorefp}[1]{(\autoref{#1})}
\newcommand{\autorefseq}[2]{\autoref{#1}--\ref{#2}}
\newcommand{\autorefseqp}[2]{(\autoref{#1}--\ref{#2})}

\DeclareMathOperator*{\aggr}{\textsc{Aggr}}
\DeclareMathOperator*{\update}{\textsc{Up}}
\DeclareMathOperator*{\msg}{\textsc{Msg}}

\DeclareMathOperator*{\sumaggr}{\textsc{Sum}}

\newcommand{\mlp}{\text{MLP}}

\newcommand{\tcn}{\text{TCN}}
\newcommand{\mpl}{\text{MP}}
\newcommand{\stmpl}{\text{STMP}}

\def\bigO{\mathcal{O}}

\newcommand{\newterm}[1]{{\emph{#1}}}

\def\eqref#1{equation~\ref{#1}}
\def\1{\bm{1}}

\def\vzero{{\bm{0}}}
\def\vone{{\bm{1}}}

\def\vtheta{{\bm{\theta}}}
\def\vomega{{\bm{\omega}}}
\def\va{{\bm{a}}}
\def\vb{{\bm{b}}}

\def\ve{{\bm{e}}}

\def\vh{{\bm{h}}}

\def\vm{{\bm{m}}}

\def\vq{{\bm{q}}}
\def\vr{{\bm{r}}}

\def\vu{{\bm{u}}}
\def\vv{{\bm{v}}}

\def\vx{{\bm{x}}}

\def\vz{{\bm{z}}}
\def\vomega{{\bm{\omega}}}

\def\mA{{\bm{A}}}

\def\mC{{\bm{C}}}

\def\mH{{\bm{H}}}

\def\mM{{\bm{M}}}

\def\mO{{\bm{O}}}

\def\mQ{{\bm{Q}}}
\def\mR{{\bm{R}}}

\def\mU{{\bm{U}}}
\def\mV{{\bm{V}}}
\def\mW{{\bm{W}}}
\def\mX{{\bm{X}}}

\def\mZ{{\bm{Z}}}

\def\mPhi{{\bm{\Phi}}}

\DeclareMathAlphabet{\mathsfit}{\encodingdefault}{\sfdefault}{m}{sl}
\SetMathAlphabet{\mathsfit}{bold}{\encodingdefault}{\sfdefault}{bx}{n}

\def\gD{{\mathcal{D}}}
\def\gE{{\mathcal{E}}}

\def\gG{{\mathcal{G}}}

\def\gN{{\mathcal{N}}}

\def\sI{{\mathbb{I}}}

\def\sR{{\mathbb{R}}}

\newcommand{\E}{\mathbb{E}}

\DeclareMathOperator*{\argmin}{arg\,min}

\DeclareMathOperator{\sign}{sgn}

\newcommand{\bigo}{\mathcal{O}}

\begin{document}

\title{Graph Deep Learning for Time Series Forecasting}

\author{Andrea Cini}
\email{andrea.cini@usi.ch}
\affiliation{%
  \institution{Universit\`a della Svizzera italiana, IDSIA}
  \city{Lugano}
  \country{Switzerland}
}

\author{Ivan Marisca}
\email{ivan.marisca@usi.ch}
\affiliation{%
  \institution{Universit\`a della Svizzera italiana, IDSIA}
  \city{Lugano}
  \country{Switzerland}
}

\author{Daniele Zambon}
\email{danile.zambon@usi.ch}
\affiliation{%
  \institution{Universit\`a della Svizzera italiana, IDSIA}
  \city{Lugano}
  \country{Switzerland}
}

\author{Cesare Alippi}
\email{cesare.alippi@usi.ch}
\affiliation{%
  \institution{Universit\`a della Svizzera italiana, IDSIA}
  \city{Lugano}
  \country{Switzerland}
}
\affiliation{%
  \institution{Politecnico di Milano}
  \city{Milan}
  \country{Italy}
}

\thanks{Published as a tutorial paper in \textit{ACM Computing Surveys}: \url{https://doi.org/10.1145/3742784}. This work was partly supported by the Swiss National Science Foundation grants No.~204061~(\emph{HORD GNN: Higher-Order Relations and Dynamics in Graph Neural Networks}) and No.~225351~(\emph{Relational Deep Learning for Reliable Time Series Forecasting at Scale}).}

\renewcommand{\shortauthors}{A.\ Cini et al.}


\begin{abstract}
Graph deep learning methods have become popular tools to process collections of correlated time series. Unlike traditional multivariate forecasting methods, graph-based predictors leverage pairwise relationships by conditioning forecasts on graphs spanning the time series collection. The conditioning takes the form of architectural inductive biases on the forecasting architecture, resulting in a family of models called spatiotemporal graph neural networks. These biases allow for training global forecasting models on large collections of time series while localizing predictions w.r.t.\ each element in the set~(nodes) by accounting for correlations among them~(edges). Recent advances in graph neural networks and deep learning for time series forecasting make the adoption of such processing framework appealing and timely. However, most studies focus on refining existing architectures by exploiting modern deep-learning practices. Conversely, foundational and methodological aspects have not been subject to systematic investigation. To fill this void, this tutorial paper aims to introduce a comprehensive methodological framework formalizing the forecasting problem and providing design principles for graph-based predictors, as well as methods to assess their performance. In addition, together with an overview of the field, we provide design guidelines and best practices, as well as an in-depth discussion of open challenges and future directions.
\end{abstract}

%    <concept>
%        <concept_id>10010147.10010257</concept_id>
%        <concept_desc>Computing methodologies~Machine learning</concept_desc>
%        <concept_significance>500</concept_significance>
%        </concept>
%    <concept>
%        <concept_id>10010147.10010257.10010293.10010294</concept_id>
%        <concept_desc>Computing methodologies~Neural networks</concept_desc>
%        <concept_significance>500</concept_significance>
%        </concept>
%    <concept>
%        <concept_id>10010147.10010178</concept_id>
%        <concept_desc>Computing methodologies~Artificial intelligence</concept_desc>
%        <concept_significance>500</concept_significance>
%        </concept>
%  </ccs2012>

% \ccsdesc[500]{Computing methodologies~Machine learning}
% \ccsdesc[500]{Computing methodologies~Neural networks}
% \ccsdesc[500]{Computing methodologies~Artificial intelligence}

\keywords{time series forecasting, graph deep learning, graph neural networks}
% \\[.3em]DOI: \url{https://doi.org/10.1145/3742784}

\maketitle

\section{Introduction}

Shallow and deep neural architectures have been used to forecast time series for decades resulting in stories of both failure~\citep{zhang1998forecasting} and success~\citep{smyl2020hybrid, kunz2023deep}. One of the key elements enabling most of the recent achievements in the field is the training of a single global neural network -- with shared parameters -- on large collections of related time series~\citep{salinas2020deepar, benidis2022deep}. Indeed, training a single \textit{global} model allows for scaling the complexity of the architecture given the larger available sample size. Such an approach, however, considers each time series independently from the others and, as a consequence, does not take into account dependencies that might be instrumental for accurate predictions~\citep{granger1969investigating}. For example, the large variety of sensors that permeates modern cyber-physical infrastructures~(e.g., traffic networks and smart grids) produces sets of time series with inherently rich \textit{spatiotemporal structure} and \textit{spatiotemporal dynamics}. On one hand, global models appear inadequate in capturing such dependencies across time series, while,  on the other, training a single \textit{local} predictor, i.e., modeling the full collection as a large multivariate time series, would negate the benefits brought by sharing the trainable parameters. Furthermore, both approaches would not allow for exploiting any prior information, such as the directionality and sparsity of the dependencies. 

The way out, as it often happens with major advancements in both deep learning~\citep{lecun1998convolutional, hochreiter1997long, scarselli2008graph} and time series forecasting~\citep{harvey1990forecasting, durbin2012time}, is to consider the structure of the data as an inductive bias. Indeed, dependencies can be represented in terms of pairwise relationships among the time series in the collection. The resulting representation is a graph where each time series is associated with a node and functional dependencies among them are represented as edges. The conditioning of the predictor on observations at correlated time series can then take the form of an architectural inductive bias in the processing carried out by the neural architecture. \Glspl{gnn}~\citep{bacciu2020gentle, bronstein2021geometric}, based on the \gls{mp} framework~\citep{gilmer2017neural}, provide the suitable neural operators allowing for sharing parameters in the processing of the time series, while, at the same time, conditioning the predictions w.r.t.\ observations at neighboring nodes~(related time series). The resulting models, operating over both time and space, are known as \glspl{stgnn}~\citep{seo2018structured, li2018diffusion, cini2023taming, jin2023survey}. \glspl{stgnn} implement global and inductive architectures for time series processing by exploiting the \gls{mp} mechanisms to account for spatial -- other than temporal -- dynamics, with the term \textit{spatial} referring to dynamics that span the collection across different time series.

Researchers have been proposing a large variety of \glspl{stgnn} by integrating \gls{mp} into popular architectures, e.g., by exploiting \gls{mp} blocks to implement the gates of recurrent cells~\citep{seo2018structured, li2018diffusion, cini2022filling, micheli2022discrete} and to propagate representations in fully convolutional~\citep{yu2018spatio, wu2019graph} and attention-based architectures~\citep{zheng2020gman, wu2021traversenet, marisca2022learning}. The adoption of the resulting \glspl{stgnn} has been successful in a wide range of time series processing applications ranging from traffic flow prediction~\citep{li2018diffusion, yu2018spatio, wu2019graph} and air quality monitoring~\citep{chen2021group, iskandaryan2023graph} to energy analytics~\citep{eandi2022spatio, cini2023scalable}, financial time series processing~\citep{chen2018incorporating, matsunaga2019exploring} and epidemiological data analysis~\citep{kapoor2020examining, fritz2022combining}. However, despite the rich literature on architectures and successful applications, the methodological foundations of the field have not been systematically laid out yet. For instance, the categorization of \glspl{stgnn} as global models and the interplay between globality and locality have been studied in such context only recently~\citep{cini2023taming}, regardless of the profound practical implications. We argue that a comprehensive formalization and methodological framework for the design of graph-based deep learning methods in time series forecasting is missing. 
The goal of this paper is, hence, to frame the problem from the proper perspective and propose a framework instrumental to tackling the inherent challenges of the field, ranging from learning the latent graph underlying the observed data to dealing with local effects, missing data, and scalability issues.

Our contributions can be summarised as follows. We
\begin{itemize}
    \item provide a formalization of the problem settings~\autorefp{sec:problem} and of time series forecasting given relational side information and the inductive biases associated with the proposed graph-based framework~\autorefp{sec:forecasting}; 
    \item present guidelines to design effective graph-based forecasting architectures \autorefseqp{sec:stgnns}{sec:global-local} and to evaluate their performance~\autorefp{sec:test};
    \item identify the challenges inherent to such problem settings and discuss the associated design choices~\autorefp{sec:challenges} addressing, in particular, the problem of dealing with missing data~\autorefp{sec:imputation}, latent graph learning~\autorefp{sec:graph-learning}, the scalability of the resulting models~\autorefp{sec:scalability} and learning in inductive settings~\autorefp{sec:inductive}.
\end{itemize}
Simulation results~\autorefp{sec:experiments} and a discussion of the related works~\autorefp{sec:related} and future directions~\autorefp{sec:future} complete the paper. We believe that the introduced comprehensive design framework will aid researchers in investigating the foundational aspects of graph deep learning for time series processing. At the same time, the paper offers a tutorial to the practitioner, providing the practical and theoretical guidelines needed to apply the introduced methodologies to real-world problems. 

\section{Related works}\label{sec:related}

Graph deep learning methods have found widespread application in the processing of temporal data~\citep{kazemi2020representation, zambon2022phdthesis, gravina2023deep, longa2023graph, jin2023survey, chen2023graph}. In this section, we review previous related works that investigate different sub-areas within the field.

\paragraph{Dynamic relational data} 
The term \emph{temporal graph} (or temporal network) is used to indicate scenarios where nodes, attributes, and edges of a graph are dynamic and are given over time as a sequence of events localized at specific nodes and/or as the interactions among them~\citep{kazemi2020representation, longa2023graph}. A typical reference application is the processing of the dynamic relationships and user profiles that characterize social networks and recommender systems. \citet{kazemi2020representation} propose an encoder-decoder framework to unify existing representation learning methods for dynamic graphs. \citet{barros2021survey} compiled a rich survey of methods for embedding dynamic networks, while \citet{skarding2021foundations} focus on \gls{gnn} approaches to the same problem. \citet{longa2023graph} introduce a taxonomy of tasks and models in temporal graph processing; \citet{gravina2023deep}, along with a categorization of existing architectures, introduce a benchmark based on a diverse set of available datasets. \citet{huang2023temporal} build an alternative set of benchmarks and datasets with a focus on applications to large-scale temporal graphs. Besides temporal graphs, a large body of literature has been dedicated to the processing of sequences of arbitrary graphs, e.g., without assuming any correspondence between nodes across time steps~\citep{zambon2022phdthesis, zambon2018concept}. Although the settings we deal with in this paper could formally be seen as a sub-area of temporal graph processing, having actual time series associated with each node radically changes the approach to the problem, as well as the available model designs and target applications. Indeed, none of the above-mentioned frameworks explicitly target time series forecasting. 

\paragraph{Graph-based time series processing} 
\Acrlongpl{stgnn} for time series processing have been pioneered in the context of traffic forecasting~\citep{li2018diffusion, yu2018spatio} and the application of graph deep learning methods in traffic analytics have been extremely successful~\citep{ye2020build, jiang2022graph, jin2023spatio}. The analysis of \glspl{stgnn} in the context of global and local forecasting models has been initiated in~\citep{cini2023taming}. \citet{jin2023survey} and \citet{chen2023graph} carried out an in-depth survey of \glspl{gnn} architectures for time series forecasting, classification, imputation, and anomaly detection. In contrast, the present paper does not focus on surveying architectures but on providing a methodological framework and a tutorial. More similarly in spirit to our work, \citet{benidis2022deep} offer a tutorial and a critical discussion of modern practices in deep learning for time series forecasting. Analogously, \citet{bronstein2021geometric} and \citet{bacciu2020gentle} provide frameworks for understanding and developing graph deep learning methods. Finally, outside of deep learning, graph-based methods for time series processing have been studied in the context of graph signal processing~\citep{ortega2018graph, stankovic2020gsp, leus2023graph} and go under the name of \textit{time-vertex signal processing} methods~\citep{grassi2017time}.

\section{Problem settings}\label{sec:problem}

This section formalizes the problem settings. In particular, \autoref{sec:settings} introduces the reference framework, suitably extended in \autoref{sec:extension} to deal with specific scenarios typical of various application domains.

\subsection{Reference problem settings}\label{sec:settings}

\begin{figure}
    \centering
    \includegraphics[width=\textwidth]{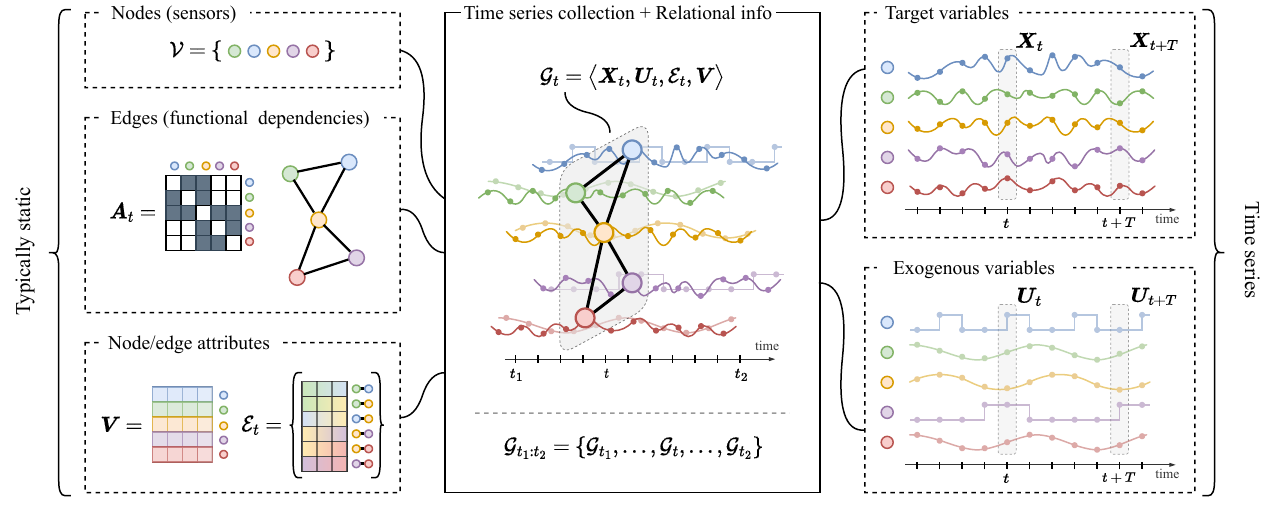}
    \caption{A collection of synchronous and regularly sampled time series with associated pairwise dependencies.}
    \label{fig:time-series}
\end{figure}

We consider collections of correlated time series with side relational information.

\paragraph{Correlated time series} Consider a collection of $N$, regularly and synchronously sampled, correlated time series;  the $i$-th time series of the collection is composed by a sequence of $d_x$-dimensional vectors $\vx^i_t \in \sR^{d_x}$ observed at each time step $t$ and coming from sensors\footnote{Here the term sensor has to be considered in a broad sense, as an entity producing a sequence of observations over time.} with  $d_x$ channels each. All $N$ time series are assumed to be \newterm{homogenous}, i.e., characterized by the same variables~(observables) -- say the same $d_x$ channels. Matrix $\mX_t \in \sR^{N\times d_x}$ denotes the stacked $N$ observations at time $t$, while  $\mX_{t:t+T}$ indicates the sequence of observations within time interval $[t, t+T)$; with the shorthand $\mX_{<t}$ we indicate observations at time steps up to $t$~(excluded). Exogenous variables associated with each time series are denoted by $\mU_t\in\sR^{N\times d_u}$, while static (time-independent) attributes are grouped in matrix $\mV \in \sR^{N\times d_v}$. We consider a setup where observations have been generated by a \newterm{time-invariant} spatiotemporal stochastic process such that
\begin{equation}
    \vx_t^i \sim p^i\left(\vx_t^i \big| \mX_{<t}, \mU_{\leq t}, \mV\right) \quad \forall i=1,\dots, N;
\end{equation}
in particular, we assume the existence of a predictive causality \textit{\`a la Granger}~\citep{granger1969investigating}, i.e., we assume that forecasts for a single time series can benefit -- in terms of accuracy -- from accounting for the past values of (a subset of) other time series in the collection.
We do not assume that the same stochastic process generates all the $N$ time series in the collection. This means that, in general, $p^i \ne p^j$ if $i \ne j$, while the assumption of time invariance remains valid. %However, we assume each process is time-invariant, i.e., fixed over time.} 
In the sequel, the term \newterm{spatial} refers to the dimension of size $N$, that spans the time series collection; in the case of physical sensors, the term spatial reflects the fact that each time series might correspond to a different physical location.

\paragraph{Relational information} Relational dependencies among the time series collection can be exploited to inform the downstream processing and allow for, e.g., getting rid of spurious correlations in the observed sequences of data. Pairwise relationships existing among the time series can be encoded by a~(possibly dynamic)  adjacency matrix $\mA_t \in \{0,1\}^{N\times N}$ that accounts for the~(possibly asymmetric) dependencies at time step $t$; optional edge attributes $\ve^{ij}_t \in \sR^{d_e}$ can be associated to each non-zero entry of $\mA_t$. In particular, we denote the set of attributed edges encoding all the available relational information by ${\gE_t \doteq \{\langle(i,j), \ve^{ij}_t\rangle\ \vert\ \forall i,j: \mA_t[i,j]\neq0\}}$. Whenever edge attributes are scalar, i.e., $d_e=1$, edge set $\gE_t$ can be simply represented as a weighted and real-valued adjacency matrix $\mA_t \in \sR^{N \times N}$. Analogously to the homogeneity assumption for observations, edges are assumed to indicate the same type of relational dependencies~(e.g., physical proximity) and have the same type of attributes across the collection. 
We use interchangeably the terms \textit{node} and \textit{sensor} to indicate each of the $N$ entities generating the time series and refer to the node set together with the relational information as \emph{sensor network}. The tuple $\gG_t \doteq \langle\mX_t, \mU_t, \gE_t, \mV\rangle$ indicates all the available information at time step $t$. Finally, note that for many applications~(e.g., traffic networks) changes in the topology happen slowly over time and the adjacency matrix -- as well as edge attributes -- can be considered as fixed within a short window of observations, i.e., $\gE_t = \gE$ and $\ve^{ij}_t = \ve^{ij}$ for all $(i,j)$ pairs. A graphical representation of the problem settings is shown in \autoref{fig:time-series}.

\begin{example} Consider a sensor network monitoring the speed of vehicles at crossroads. In this case, $\mX_{1:t}$ refers to past traffic speed measurements sampled at a certain frequency. Exogenous variables $\mU_t$ account for time-of-the-day and day-of-the-week identifiers, and the current weather conditions. The node-attribute matrix $\mV$ collects static features related to the sensor's position, e.g., the type of road the sensor is placed in or the number of lanes. A static adjacency matrix $\mA$ can be obtained by considering each pair of sensors connected by an edge -- weighted by the road distance -- if and only if a road segment directly connects them. Conversely, road closures and traffic diversions can be accounted for by adopting a dynamic topology $\mA_t$.
\end{example}

\subsection{Extensions to the reference settings}\label{sec:extension}
This section offers extensions to the reference problem settings by discussing how the framework can be modified to account for peculiarities typical of a wide range of practical applications.

\paragraph{New nodes, missing observations, and multiple collections} It is often the case that the time frames of the time series in the collection, although synchronous and regularly sampled, do not overlap perfectly, i.e., some time series might become available at a later time and there might be windows with blocks of missing observations. For example, it is typical for the number of installed sensors to grow over time and many applications are affected by the presence of missing data, e.g., associated with readout and/or communication failures which result in transient or permanent faults. These scenarios can be incorporated into the framework by setting $N$ to the total maximum number of time series available, and, whenever needed, padding the time series appropriately to allow for a tabular representation of $\{\mX_t\}_t$. An auxiliary binary exogenous variable $\mM_t \in \{0,1\}^{N \times d_x}$, called \textit{mask}, can be introduced at each time step as $\gG_t \doteq \langle\mX_t, \mU_t, \mM_t, \gE_t, \mV\rangle$ to model the availability of observations w.r.t.\ each node and time step. In particular, we set $\vm^i_t{\scriptstyle[k]} = 1$ if $k$-th channel in the corresponding observation $\vx^i_t$ is valid, and $\vm^i_t{\scriptstyle[k]} = 0$ otherwise. If observations associated with the $i$-th node are completely missing at time step $t$, the associated mask vector will be null, i.e.,  $\vm^i_t = \vzero$. The masked representation simplifies the presentation of concepts and, at the same time, is useful in data reconstruction tasks~(see \autoref{sec:imputation}).  Finally, if collections from multiple sensor networks are available, the problem can be formalized as learning from $M$ disjoint sets of correlated time series $\gD = \big\{\gG^{(1)}_{t_1:t_1+T_1},\gG^{(2)}_{t_2:t_2+T_2}, \dots, \gG^{(M)}_{t_m:t_m+T_m}\big\}$, potentially without overlapping time frames. In the latter case, we assume the absence of functional dependencies between time series in different sets and the homogeneity of node features and edge attributes across collections.

\paragraph{Heterogeneous time series and edge attributes} Heterogeneous sets of correlated time series are commonly found in the real world~(e.g., consider a set of weather stations equipped with different sensory packages) and result in collections where observations across the time series in the set might correspond to different variables. Luckily, dealing with this setting is relatively straightforward and can be done in several ways. In particular, the masked representation introduced in the above paragraph can be used to pad each time series to the same dimension $d_{max}$ and keep track of the available channels at each node; moreover, the sensor type of each sensor can be encoded in the attribute matrix $\mV$. If the total number of variables is too large or is expected to change over time, one alternative strategy is to map each observation into a shared homogenous representation~(see, e.g., relational models such as \citep{schlichtkrull2018modeling}). Heterogeneous edge attributes can be dealt with analogously to heterogeneous node features.

\section{Graph-based Time Series Forecasting}\label{sec:forecasting}

We address in the following the multi-step-ahead time-series forecasting problem~\citep{benidis2022deep}, i.e., we are interested in predicting, for each time step $t$ and some forecasting horizon $H\ge 1$, the $h$ step-ahead observations $\mX_{t+h}$ for all $h \in [0, H)$ given a window of $W\ge 1$ past observations. In particular, we are interested in learning a model $p_\vtheta$ approximating the unknown conditional probability
\begin{equation}\label{eq:fc-model}
    p_\vtheta\left(\vx_{t+h}^i \big| \mX_{t-W:t}, \mU_{t-W:t+h+1}, \mV\right) \approx p^i\left(\vx_{t+h}^i \big| \mX_{<t}, \mU_{\leq t+h}, \mV\right) \quad \forall h \in [0, H), \forall i=1,\dots,N
\end{equation}
where $\vtheta$ indicates the learnable parameters of the model which may or may not be specialized w.r.t.\ the $i$-th time series~(see \autoref{sec:global-local}). 
Note that not all the exogenous variables $\mU_{\leq t+h}$ might be available up to time step $t+h$ in practical applications%
\footnote{Exogenous variables might contain, for example, actual weather conditions~(available up to time step $t$) or estimated values, e.g., weather forecasts, available for future time steps as well (up to time step ${t+h}$).}; in such cases, predictions will be conditioned on covariates up to $t-1$, i.e., $\mU_{t-W:t}$. 

\paragraph{Relational Inductive Biases for Time Series Forecasting} Learning an accurate model $p_{\vtheta}$ following \autoref{eq:fc-model} can become increasingly difficult as the number of time series in the collection grows. Intuitively, the high dimensionality of the problem can lead to spurious correlations among the observed time series, impairing the effectiveness of the learning procedure. One way to address this issue is to embed the available relational information as an inductive bias into the model. In particular, dependencies among the time series can be used to condition the prediction and, as discussed in \autoref{sec:stgnns}, accounted for in the predictor through an architectural bias. The considered family of models can then be written as
\begin{equation}\label{eq:graph-based-prob-model}
    p_\vtheta\left(\vx_{t+h}^i \big| \gG_{t-W:t}, \mU_{t:t+h+1}\right) \approx p^i\left(\vx_{t+h}^i \big| \mX_{< t}, \mU_{\leq t+h}, \mV\right) \quad \forall h \in [0, H).
\end{equation}
Notably, the conditioning on the sequence of attributed graphs $\gG_{t-W:t}$ and, in particular, on the relationships encoded in $\gE_{t-W:t}$, can localize predictions w.r.t.\ the neighborhood of each node and is intended to constrain the model to the most plausible ones. In the sequel, we focus on point forecasts, i.e., we limit our analysis to the problem of predicting point estimates rather than modeling a full probability distribution. Under such assumption, we can consider predictive model families $\mathcal{F}({}\cdot{};\vtheta)$ such that
\begin{equation}\label{eq:graph-based-predictor}
    \widehat{\mX}_{t:t+H} = \mathcal{F}\left(\gG_{t-W:t}, \mU_{t:t+h+1}; \vtheta\right)\quad \text{s.t.} \quad \widehat{\mX}_{t:t+H} \approx \E_p\left[\mX_{t:t+H}\right].
\end{equation}
Parameters $\vtheta$ can be learned by minimizing a cost function $\ell({}\cdot{})$ on a training set, i.e., 
\begin{align}\label{eq:loss}
    \widehat\vtheta &= \argmin_\vtheta \frac{1}{T}\sum_{t=1}^{T}\ell\left(\widehat \mX_{t:t+H}, \mX_{t:t+H} \right) ,
\end{align}
where the cost is, e.g., the \textit{squared error} 
\begin{align}\label{eq:l2-loss}
   \ell\left(\widehat \mX_{t:t+H}, \mX_{t:t+H} \right) &= \frac{1}{N\,H}\sum_{i=1}^N\sum_{h=0}^{H-1}\Big\| \widehat \vx^i_{t+h} - \vx^i_{t+h} \Big\|_2^2.
\end{align}
The following sections delve into the design of $\mathcal{F}({}\cdot{}; \vtheta)$ and graph deep learning methods to embed relational inductive biases~\citep{battaglia2018relational} into the processing architecture.

\section{Spatiotemporal Graph Neural Networks}\label{sec:stgnns}

This section introduces \gls{mp} operators and their use in deep neural network architectures to process multiple time series; the framework follows \citet{cini2023taming}. As already discussed, architectures within this framework are usually referred to as \glspl{stgnn}~\citep{li2018diffusion, yu2018spatio}. \glspl{stgnn} are global forecasting models where parameters are shared among the target time series; the discussion on this fundamental aspect -- already mentioned in the introduction -- and on hybrid global-local architectures will be resumed in \autoref{sec:global-local}.  Although the focus of the paper is not on providing a taxonomy of the existing architectures, we discuss in this section the design choices available to the practitioner; we refer to \citet{jin2023survey} for an in-depth survey on the existing architecture across different tasks in time series processing.

\subsection{Message-passing neural networks}

Modern \glspl{gnn}~\citep{scarselli2008graph, bacciu2020gentle, bronstein2021geometric} embed architectural biases into the processing architecture by constraining the propagation of information w.r.t. a notion of neighborhood derived from the adjacency matrix. Most of the commonly used architectures fit into the \gls{mp} framework~\citep{gilmer2017neural}, which provides a recipe for designing \gls{gnn} layers; \glspl{gnn} that fit within the \gls{mp} framework are usually referred to as \textit{spatial \glspl{gnn}}, usually in opposition to \textit{spectral \glspl{gnn}}, which instead operate in the spectral domain\footnote{Note that most of the so-called spectral \glspl{gnn} can be seen as special instances of \gls{mp} architectures nonetheless.}~\citep{bruna2014spectral, wang2022powerful}. By taking as reference a graph with static node features $\mH^0 \in \sR^{N\times d_h}$ and edge set $\gE$, we consider \gls{mp} neural networks obtained by stacking \gls{mp} layers that update each $i$-th node representation at each $l$-th layer as
\begin{equation}
    \vh^{i,l+1} = {\update}^l{\Big(\vh^{i,l}, \aggr_{j \in \mathcal{N}(i)}}\Big\{{\msg}^l{\big(\vh^{i,l}, \vh^{j,l},\ve^{ji}\big)}\Big\}\Big),\label{eq:mp}\\
\end{equation}
where ${\update}^l({}\cdot{})$ and ${\msg}^l({}\cdot{})$ are respectively the update and message functions, e.g., implemented by \glspl{mlp}. $\textsc{Aggr}\{{}\cdot{}\}$ indicates a generic permutation invariant aggregation function, while $\mathcal{N}(i)$ refers to the set of neighbors of node $i$, each associated with edge attribute $\ve^{ji}$. In the following, we use the shorthand ${\mH^{l+1} = \mpl^l\left(\mH^l, \gE\right)}$ to indicate a \gls{mp} step w.r.t.\ the full node set. \gls{mp} \glspl{gnn} are \textit{inductive} models~\citep{ruiz2020graphon} which can process unseen graphs of variable sizes by sharing weights among nodes and localizing representations by aggregating features at neighboring nodes. 

By following~\citet{dwivedi2020benchmarking}, we call \textit{isotropic} those \glspl{gnn} where the message function ${\msg}^l$ only depends on the features of the sender node $\vh^{j,l}$; conversely, we use the term \textit{anistropic} referring to GNNs where ${\msg}^l$ takes both $\vh^{i,l}$ and $\vh^{j,l}$ as input. For instance, a standard and commonly used isotropic \gls{mp} layer for weighted graphs~(with weighted adjacency matrix $\mA$) is
\begin{equation}
    \vh^{i,l+1} = \xi\Big(\mW_1^l \vh^{i,l} + \sumaggr_{j \in \mathcal{N}(i)} \Big\{a^{ji}\mW_2^l \vh^{j,l}\Big\} \Big), \label{eq:imp}
\end{equation}
where $\mW_1^l$ and $\mW_2^l$ are matrices of learnable parameters, $a^{ji} = \mA[j,i]$, and $\xi({}\cdot{})$ is a generic activation function. Conversely, an example of an anisotropic \gls{mp} operator, based on~\citep{bresson2017residual}, is 
\begin{align}
    \vm^{j\rightarrow i, l} &= \mW_2^l \xi\Big(\mW_1^l\Big[\vh^{i,l}|| \vh^{j,l}|| \ve^{ji}\Big]\Big), \label{eq:amp_first}\hspace{2em}
    \alpha^{ji, l} = \sigma\Big(\mW_0^l\vm^{ji, l}\Big),\\
    \vh^{i,l+1} &= \xi\Big(\mW^l_3 \vh^{i,l} + \sumaggr_{j \in \mathcal{N}(i)}  \Big\{\alpha^{ji, l} \vm^{j\rightarrow i, l}\Big\} \Big), \label{eq:amp_last}
\end{align}
where matrices ${\mW_0^l\in\sR^{1\times d_m}}$, $\mW^l_1$, $\mW^l_2$ and $\mW^l_3$ are learnable parameters, $\sigma({}\cdot{})$ is the sigmoid activation function and $||$ the concatenation operator applied along the feature dimension. 
Intuitively, isotropic \gls{mp} operators compute and aggregate messages without taking into account the representations of sender and receiver nodes and rely entirely on the presence of edge weights to weigh the contribution of different neighbors. Conversely, anisotropic schemes allow for learning adaptive aggregation and message-passing schemes aware of the nodes involved in the computation. Popular anisotropic operators exploit multi-head attention mechanisms to learn rich propagations schemes where the information flowing from each neighbor is weighted and aggregated after multiple parallel transformations~\citep{velivckovic2018graph, vaswani2017attention}. Indeed, we point out that selecting the proper \gls{mp} operator, i.e., choosing the architectural bias for constraining the flow of information, is crucial for obtaining good performance for the problem at hand. In fact, standard isotropic filters are often based on homophily -- i.e., the assumption that neighboring nodes behave in a similar way -- and can suffer from over-smoothing~\citep{rusch2023survey}.

\subsection[Spatiotemporal message-passing]{\Acrlong{stmp}}
By following the terminology introduced in~\citep{cini2023taming}, \glspl{stgnn} can be designed by extending \gls{mp} to aggregate, at each time step, spatiotemporal information from each node's neighborhood; in particular, a \gls{stmp} block updates representations as
\begin{equation}
    \vh^{i,l+1}_{t} = {\update}^l\Big(\vh^{i,l}_{\leq t}, \aggr_{j \in \mathcal{N}_t(i)}\Big\{\textsc{Msg}^l\big(\vh^{i,l}_{\leq t}, \vh^{j,l}_{\leq t},\ve^{ji}_{\leq t}\big)\Big\}\Big),\label{eq:stmp}\\
\end{equation}
where $\mathcal{N}_t(i)$ indicates the neighbors of the $i$-th node at time step $t$~(i.e., the nodes associated with incoming edges in $\gE_t$). As in the previous case, in the following, the shorthand $\smash{\mH^{l+1}_{t} = \stmpl^l\big(\mH^{l}_{\leq t}, \gE_{\leq t}\big)}$ indicates an \gls{stmp} step. Blocks of an \gls{stmp} layer will have to be designed, then, to handle sequences of data.  
The next section provides recipes for building \glspl{stgnn} based on different implementations of the \gls{stmp} blocks and on existing popular \gls{stgnn} architectures.

\section{Forecasting architectures}\label{sec:stgnn-architectures}

We consider forecasting architectures consisting of an encoding step followed by \gls{stmp} layers and a final readout mapping representations to predictions. As such, models introduced in \autoref{eq:graph-based-predictor} can be framed as a sequence of three operations performed at each time step: 
\begin{align}
    \vh^{i,0}_{t-1} &= \textsc{Encoder}\left(\vx^{i}_{t-1}, \vu^{i}_{t-1}, \vv^i\right),\label{eq:enc}\\
    \mH^{l+1}_{t-1} &= \textsc{STMP}^l\Big(\mH^{l}_{\leq t-1}, \gE_{\leq t-1}\Big),\quad l=0,\dots,L-1\label{eq:stmp-block}\\
    \hat \vx^{i}_{t:t+H} &= \textsc{Decoder}\left(\vh_{t-1}^{i,L}, \vu^i_{t:t+H}\right)\label{eq:dec}.
\end{align}
$\textsc{Encoder}({}\cdot{})$ and $\textsc{Decoder}({}\cdot{})$ indicate generic encoder and readout layers that can be implemented, as an example, as standard fully connected linear layers, or \glspl{mlp}. Note that both encoder and decoder do not propagate information along time and space. By adopting the terminology of previous works~\citep{gao2021equivalence, cini2023sparse, cini2023taming}, we categorize \glspl{stgnn} following this scheme in \gls{tts}, \gls{stt}, and \gls{tas} models. More specifically, in a \gls{tts} model the sequence of representations $\vh^{i,0}_{< t}$ is processed by a sequence model, such as a \gls{rnn}, before any \gls{mp} operation along the spatial dimension~\citep{gao2021equivalence}; \gls{stt} models are similarly obtained by inverting the order of the two operations. Conversely, in \gls{tas} models time and space are processed in a more integrated fashion, e.g., by a recurrent \gls{gnn}~\citep{seo2018structured} or by spatiotemporal convolutional operators~\citep{yu2018spatio}. 

\paragraph{\Acrlong{tas} models} We include in this category any \gls{stgnn} in which the processing of the temporal and spatial dimensions cannot be factorized in two separate steps. In \gls{tas} models, representations at every node and time step are the result of joint temporal and spatial processing as in \autoref{eq:stmp-block}. 
To the best of our knowledge, the first \gls{tas} \glspl{stgnn} have been proposed by \citet{seo2018structured}, who introduced a popular family of recurrent architectures, hereby denoted as \glspl{gcrnn}, where standard fully-connected layers in~(gated) \glspl{rnn} are replaced by graph convolutions~\citep{kipf2017semi, defferrard2016convolutional}. As an example, by considering a \gls{gru} cell~\citep{cho2014properties} and replacing graph convolutions with generic \gls{mp} layers, the resulting recurrent model updates representations at each time step $t$ as
\begin{align}
    \mZ_t^l &= \mH_t^{l-1}\\
    \mR_t^{l} &= \sigma\left(\mpl^l_r\left(\left[\mZ_t^{l} || \mH_{t-1}^{l} \right], \gE_t \right)\right),\label{eq:gcrnn-first} \\
    \mO_t^{l} &= \sigma\left(\mpl^l_o\left(\left[ \mZ_t^{l} || \mH_{t-1}^{l} \right], \gE_t \right)\right), \\
    \mC_t^{l} &= \text{tanh}\left(\mpl^l_c\left(\left[ \mZ_t^{l} || \mR_t^{l} \odot \mH_{t-1}^{l} \right], \gE_t \right)\right), \\
    \mH_t^{l} &= \mO_t^{l} \odot \mH_{t-1}^{l} + (1 - \mO_t^{l}) \odot \mC_t^{l}\label{eq:gcrnn-last},
\end{align}
with ${}\odot{}$ denoting the element-wise~(Hadamard) product and ${}||{}$ the concatenation operation. Note that we consider, for each gate, a single \gls{mp} operation at each $l$-th layer for conciseness' sake~(a stack of \gls{mp} layers is often adopted in practice). Models following a similar approach have found widespread adoption replacing standard \glspl{rnn} in the context of correlated time series processing~\citep{li2018diffusion,  zhang2018gaan, bai2020adaptive, cini2022filling}.
Apart from \glspl{gcrnn}, an approach to building \gls{tas} models consists of integrating a temporal operator directly into the $\msg({}\cdot{})$ function. Among the others, \citet{wu2021traversenet} and \citet{marisca2022learning} use cross-node attention as a mechanism to propagate information among sequences of observations at neighboring nodes. As an additional example, an analogous model could be obtained by implementing the $\update({}\cdot{})$ and $\msg({}\cdot{})$ functions of the \gls{stmp} layer in \autoref{eq:stmp} as \glspl{tcn}~\citep{bai2018empirical}:
\begin{equation}
    \vh_{t-W:t}^{i,l} = \tcn_1^l\left(\vh_{t-W:t}^{i,l-1}, \aggr_{j \in \mathcal{N}_t(i)}\left\{\tcn_2^l\left(\vh_{t-W:t}^{i,l-1}, \vh_{t-W:t}^{j,l-1}, \ve^{ji}_{t-W:t} \right)\right\}\right)\label{eq:stcn-mp}.
\end{equation}
Note that the operator resulting from the \gls{mp} processing defined in \autoref{eq:stcn-mp} can be seen as operating on the product graph obtained from spatial and temporal relationships~\citep{sabbaqi2022graph}. Finally, a straightforward approach to build \gls{tas} architectures is that of stacking blocks of alternating spatial and temporal  operators~\citep{yu2018spatio, wu2019graph, wu2020connecting}, e.g.,
\begin{equation}
    \vz^{i, l}_{t-W: t} = \tcn^l\left(\vh^{i, l-1}_{t-W:t}\right) \quad \forall i, \qquad \mH^{l}_t = \mpl^l\left(\mZ_t^{l}, \gE_t \right) \quad \forall\ t,
\end{equation}
where $\tcn^l({}\cdot{})$ indicates a temporal convolutional network layer. The first example of such architecture was introduced in~\citep{yu2018spatio}. One of the major drawbacks of \gls{tas} models is their time and space complexity which usually scale with the number of nodes and edges in the graph times the number of input time steps, i.e., with $\bigO\big(W \big(N + L|\gE_{\max}|\big)\big)$, where $N \ll|\gE_{\max}| = \max \{|\gE_{t-k}|\}^{W}_{k=1}$~(see \autoref{sec:scalability}).

\paragraph{\Acrlong{tts} models} The general recipe for a \gls{tts} model consists in 1) encoding time series associated with each node into a vector, obtaining an attributed graph, and 2)~propagating the obtained representations throughout the graph with a stack of standard \gls{mp} layers, i.e.,
\begin{align}
    &&\vh^{i,1}_t &= \textsc{SeqEnc}\left(\vh^{i,0}_{\leq t}\right),&&\label{eq:tts-seq-enc}\\
    &&{{\mH}^{l+1}_{t}} &= \mpl^{l}\left(\mH^{l}_{t}, \gE_t\right), & \forall\ l &= 1,\dots,L-1.\label{eq:tts-mp}
\end{align}
The sequence encoder $\textsc{SeqEnc}\left({}\cdot{}\right)$ can be implemented by any modern deep learning architecture for sequence modeling such as \glspl{rnn}, \glspl{tcn} and attention-based models~\citep{vaswani2017attention, li2019enhancing}. Note that this temporal encoder can consist of multiple layers, i.e., it can be a deep network by itself. Since \gls{mp} is performed only w.r.t.\ representations corresponding to the last time step, in case of a dynamic topology the edge set used for propagation can be obtained as a function of $\gE_{t-W:t}$ rather than simply using $\gE_t$, i.e., $\widetilde \gE_t = \aggr\{\gE_{t-W:t}\}$. A possible choice would be to take the union of all the edge sets, which, however, requires further processing in the case of attributed edges~\citep{gao2021equivalence}. 
\gls{tts} models are relatively uncommon in the literature~\citep{gao2021equivalence, satorras2022multivariate, cini2023scalable, cini2023graph} but are becoming more popular due to their efficiency and scalability compared to \gls{tas} alternatives~\citep{gao2021equivalence}, as discussed in \autoref{sec:scalability}. Differently from generic \gls{tas} models, in fact, the number of \gls{mp} operations does not depend on the size of the window $W$. Indeed, \gls{tts} models have a time and space complexity that scales as $\bigo\big(NW + L|\gE_t|\big)$, rather than $\bigO\big(W \big(N + L|\gE_{\max}|\big)\big)$ of \gls{tas} models. However, the two-step encoding might introduce bottlenecks in the propagation of information.

\paragraph{\Acrlong{stt} models} \gls{stt} models can be built by simply inverting the order of \autoref{eq:tts-seq-enc} and \ref{eq:tts-mp}, i.e., by using \gls{mp} layers to process static representations at each time step, then encoded along the temporal axis by a sequence model, i.e.,
\begin{align}
    &&\mH^{i,l}_t &= \mpl^{l}\left(\mH^{i,l-1}_t, \gE_t\right), & \forall\ l &= 1,\dots,L-1\label{eq:stt-mp}\\
    &&{{\vh}^{i,L}_{t}} &= \textsc{SeqEnc}\left(\vh^{i,L-1}_{t-W:t}\right).\label{eq:sst-seq-enc}
\end{align}
The general idea behind \gls{stt} approaches is to first enrich node observations by accounting for observations at neighboring nodes, and then process obtained sequences with a standard sequence model. Although they have seen some applications~\citep{seo2018structured, pareja2020evolvegcn, zhao2019t}, \gls{stt} models do not offer the same computational benefits of \gls{tts} models, having the same $\bigO\big(W \big(N + L|\gE_{\max}|\big)\big)$ complexity of \gls{tas} models. 
Nonetheless, as in \gls{tas} models, dynamic edge sets $\gE_{t-W:t}$ can be accounted for by performing \gls{mp} operations w.r.t.\ the corresponding edges at each time step. Analogously to \gls{tts} models, the factorization of the processing in two steps might introduce bottlenecks.

\section{On the Globality and Locality of Spatiotemporal Graph Neural Networks}\label{sec:global-local}

This section formally defines the concepts of globality and locality in forecasting models and emphasizes that these terms refer to model properties rather than problem settings. 
As both global and local models can be used to forecast collections of time series, the section discusses the peculiar position of \glspl{stgnn} within this context. 
Finally, hybrid global-local \gls{stgnn} architectures are introduced.

\subsection{Global and local models} 

A time series forecasting model is called \textit{global} if its parameters are fitted to a group of time series~(either univariate or multivariate); conversely, \textit{local} models are specific to a single~(possibly multivariate) time series. In different terms, a global model is trained to make predictions by learning from a set of time series, possibly generated by different stochastic processes, without learning any time-series-specific parameter. Conversely, a local model is obtained by minimizing the forecasting error on observations coming from a single time series. Forecasting multiple time series with a local model requires fitting $N$ models -- one for each target time series -- or a single~(poorly scalable) multivariate model. The advantages of global models have been discussed at length in the time series forecasting literature~\citep{salinas2020deepar, januschowski2020criteria, montero2021principles, benidis2022deep} and are mainly ascribable to the availability of large amounts of data that enable the use of models with a higher capacity w.r.t.\ single local models. Indeed, as noted by \citet{montero2021principles}, given a large enough window of observations and model complexity, if a global model is a universal function approximator it could in principle output predictions identical to those of a set of local models individual to each time series. Training a single global model increases the effective sample size available to the learning procedure and, consequently, allows for exploiting models with a higher model complexity preventing overfitting. Finally, being trained on a set of time series, global models can extrapolate to related but unseen time series, i.e., they can be used in inductive learning scenarios where target time series~(i.e., the ones to predict) can be potentially different from those in the training set\footnote{Such setting is relevant in many practical application domains and also known as the \textit{cold-start} scenario~\citep{benidis2022deep}; see \autoref{sec:inductive}.}. More formally, following \citet{benidis2022deep}, and considering our problem setting and $h$-step-ahead forecasts, a node-level local model would approximate the process generating the data as
\begin{align}\label{eq:local-node-level}p^i_{\vtheta^i}\left(\vx_{t+h}^i \big| \vx^i_{t-W:t}, \vu^i_{t-W:t+h+1}, \vv^i\right) \approx p^i\left(\vx_{t+h}^i \big| \mX_{<t}, \mU_{\leq t+h}, \mV\right) \quad i = 1, \dots, N,
\end{align}
where $\vtheta^i$ indicates the model's parameters fitted on the $i$-th time series. Differently, in a node-level global model, parameters would be shared among time series, i.e,
\begin{equation}\label{eq:global-node-level}
    p_{\vtheta}\left(\vx_{t+h}^i \big| \vx^i_{t-W:t}, \vu^i_{t-W:t+h+1}, \vv^i\right) \approx p^i\left(\vx_{t+h}^i \big| \mX_{<t}, \mU_{\leq t+h}, \mV\right) \quad i = 1, \dots, N,
\end{equation}
where parameters $\vtheta$ can be learned by minimizing the cost function w.r.t. the complete time series collection~(see \autoref{eq:loss}). The main limit of standard global models in \autoref{eq:global-node-level} is that dependencies among the synchronous time series in the collections are ignored. One option would be to consider models that simply regard the input as a single multivariate time series, i.e., with a local model such that
\begin{equation}\label{eq:local-graph-level}
    p_{\vtheta}\left(\mX_{t+h} \big| \mX_{t-W:t}, \mU_{t-W:t+h+1}, \mV\right) \approx p\left(\mX_{t+h} \big| \mX_{<t}, \mU_{\leq t+h}, \mV\right) \quad i = 1, \dots, N.
\end{equation}
However, the resulting model would not be able to exploit the advantages that come from the global perspective and would have to deal with the high dimensionality of $\mX_t$. 

\paragraph{Globality and locality in \glspl{stgnn}} \glspl{stgnn} presented in \autoref{sec:forecasting} \textit{are global models} that exploit relational architectural biases to account for related time series, going beyond the limits of the standard global approach. Indeed, by considering the \gls{stmp} scheme of \autoref{eq:stmp}, it is straightforward to see that \gls{stmp} operators share parameters among the time series in the collection and condition the representations w.r.t.\ each node's neighborhood to account for spatial dependencies that would be ignored by standard global models. \glspl{stgnn} are inductive and transferable as they do not rely upon node-specific parameters; such properties make them distinctively different from the local multivariate approach in \autoref{eq:local-graph-level}. Global models of the type implemented by \glspl{stgnn} are akin to those formalized in \autoref{eq:fc-model}, i.e.,
\begin{equation}\label{eq:global-graph-level}
    p_\vtheta\left(\vx_{t+h}^i \big| \mX_{t-W:t}, \mU_{t-W:t+h+1}, \mV\right) \approx p^i\left(\vx_{t}^i \big| \mX_{<t+h}, \mU_{\leq t+h}, \mV\right) \quad \forall i=1,\dots,N.
\end{equation}
Besides resorting to \gls{mp} operators and the relational inductive biases typical of \glspl{gnn}, global models of such class can be built by considering other classes of permutation equivariant neural operators acting on sets, such as \textit{deep sets}~\citep{zaheer2017deep} and \textit{transformers}~\citep{vaswani2017attention}; comprehensive treatment of such models is out of the scope of the present paper. As discussed in \citet{cini2023taming}, the interplay between global and local aspects plays a major role in the context of graph-based forecasting models. Indeed, although the drawbacks of the local approach are evident, global \glspl{stgnn} might struggle to account for the peculiarities of each time series in the collection and might require impractically long observation windows and large memory capacity~\citep{montero2021principles, cini2023taming}. For example, considering electric load forecasting, the consumption patterns of single residential customers are influenced not only by shared factors, e.g., weather, working hours, and holidays, but also by their individual daily routines, varying among users to different extents. By following~\citep{cini2023taming}, we refer to dynamics characterizing individual time series as \textit{local effects}. The remainder of the section discusses how to add specialized local components into otherwise global architectures to strike a balance between the global and local modeling paradigms in the context of graph-based architectures.

\subsection{Global-local \glspl{stgnn}} 

Combining global graph-based components with local node-level components has the potential for achieving a two-fold objective: 1) exploiting relational dependencies together with side information to learn flexible and efficient graph deep learning models and, at the same time, 2) obtaining accurate predictions specialized for each time series. In particular, introducing local components specific to each time series explicitly accounts for node-level effects that would not be efficiently captured by fully global models~\citep{bai2020adaptive, deng2021graph}. By doing so, the designer accepts a compromise in transferability that often empirically leads to higher forecasting accuracy on the task at hand. 
In particular, global-local \glspl{stgnn} model the data-generating process as 
\begin{equation}\label{eq:global-local-node-level}
    p^i_{\vtheta,\{\vomega^{i}\}}\left(\vx_{t+h}^i \big| \gG_{t-W:t}, \mU_{t:t+h+1}\right) \approx p^i\left(\vx_{t+h}^i \big| \mX_{<t}, \mU_{\leq t+h}, \mV\right) \quad i = 1, \dots, N,
\end{equation}
where parameter vector $\vtheta$ is shared across all nodes, whereas $\{\vomega^i\}_{i=1}^N$ are time-series dependent parameters. The associated point predictor is 
\begin{equation}
    \widehat \mX_{t:t+H} = \mathcal{F}\left(\gG_{t-W:t}, \mU_{t:t+h+1}; \vtheta, \{\vomega^i\}_{i=1}^N\right)
\end{equation}
where $\mathcal{F}({}{\cdot}{})$ is shared among all nodes.
Predictor $\mathcal{F}({}\cdot{})$ could be implemented, for example, as a sum between a global model and a~(simpler) local one:
\begin{align}
    \widehat \mX^{(1)}_{t:t+H} &= \mathcal{F}_\textsc{g}\left(\gG_{t-W:t}; \vtheta\right),&\hat \vx^{i,(2)}_{t:t+H} &= f_i\left(\vx^i_{t-W:t}; \vomega^i\right)
\end{align}
\begin{align}\label{eq:global-local-sum}
    \hat \vx^{i}_{t:t+H} &= \hat \vx^{i,(1)}_{t:t+H} + \hat \vx^{i,(2)}_{t:t+H},
\end{align}
or -- with a more integrated approach -- by using different weights for each time series at the encoding and/or decoding steps. The latter approach results in using a different encoder and/or decoder for each $i$-th node in the template \gls{stgnn} architecture \autorefseqp{eq:enc}{eq:dec} to extract representations and, eventually, project them back into the input space:
\begin{align}
    \vh^{i,0}_{t-1} &= \textsc{Encoder}_i\left(\vx^{i}_{t-1}, \vu^{i}_{t-1}, \vv^i; \vomega^i_{enc}\right),\label{eq:local-enc}\\
    \hat \vx^{i}_{t:t+H} &= \textsc{Decoder}_i\left(\vh_{t-1}^{i,L}, \vu^i_{t:t+H}; \vomega^i_{dec}\right).\label{eq:local-dec}
\end{align}
\Gls{stmp} layers could in principle be modified as well to include specialized operators, e.g., by using a different local update function $\update_i({}\cdot{})$ for each node. However, this would be impractical unless subsets of nodes are allowed to share parameters to some extent~(e.g., by clustering them~\citep{cini2023taming, cini2023graph}). Clearly, specialization compromises the use of such hybrid models in inductive learning settings \autorefp{sec:inductive} and often results in a number of learnable parameters that can be drastically higher compared to fully global models, hence compromising computational scalability as well. Associating each node to a learnable embedding provides a method to amortize the cost of specializing the model and makes transferring the learned model to different node sets easier.

\paragraph{Learnable node embeddings} The presence of static node features $\mV$ characterizing the time series in the collection might provide node identification mechanisms and, thus, alleviate the need for including specialized components in the architecture. However, such node attributes are either not available or not sufficient in most settings. Resorting to learnable node embeddings, i.e., a table of learnable parameters $\bm{\Omega} = \mQ \in \sR^{N \times d_{v}}$, offers a solution and can be interpreted as amortizing the learning of node-level specialized models~\citep{cini2023taming}. More specifically, instead of learning a local model for each time series, embeddings fed into modules of a global \gls{stgnn} can be learned end-to-end with the forecasting architecture providing a mean to condition representations at each node w.r.t.\ the peculiarities of each time series. 

The template model can be updated to account for the learned embeddings by changing the encoder and decoder as
\begin{align}
    \vh_{t-1}^{i,0} &= \textsc{Encoder}\left(\vx_{t-1}^i, \vu_{t-1}^i, \vv^i, \vq^i\right),\label{eq:node-emb-enc}\\
    \hat \vx^{i}_{t:t+H} &= \textsc{Decoder}\left(\vh_{t-1}^{i,L}, \vu^i_{t:t+H}, \vq^i\right),\label{eq:node-emb-dec}
\end{align}
which can be seen as amortized versions of the encoder and decoder in \autorefseq{eq:local-enc}{eq:local-dec}. The encoding scheme of \autoref{eq:node-emb-enc} also facilitates the propagation of relevant information by identifying nodes before message passing. \gls{stmp} layers can be updated as well to process, e.g., as an additional input of the message function, the embeddings of source and target nodes. Besides conditioning encoding and decoding steps, many of the popular \gls{stgnn} architectures use node embeddings within the processing, often as positional encodings~\citep{marisca2022learning, shao2022spatial} or to learn a factorized weighted adjacency matrix~\citep{wu2019graph, wu2020connecting, bai2020adaptive}~(see \autoref{sec:graph-learning}). Such hybrid approaches result in impressive empirical results~\citep{cini2023taming}, noticeably improving the forecasting accuracy of fully global models, and have become predominant in transductive settings, i.e., when the node set is fixed. In summary, globality and locality play a pivotal role in deep learning for time series forecasting and are aspects to consider when designing graph-based predictors. Indeed, while in practical application hybrid global-local models often outperform global architectures, trade-offs in flexibility must be taken into account.

\section{Assessing the quality of predictive models}\label{sec:test}

The quality of forecasting models is primarily assessed in terms of their performance at task. 
The best model (among many) is, then, usually selected as the one with \textit{significantly} better performance than the others~\citep{demsar2006statistical}. 
The squared error $\mathbb E\left[\lVert\vr^i_{t}\rVert_2^2\right]$
is a commonly used performance metric and is given by the 2-norm%
    \footnote{As $\mX_{t:t+H}$ is a ${H\times N\times d_x}$ tensor, norm $\lVert\mX_{t:t+H} \rVert_2^2$ is intended here as $\sum_{i=1}^N\sum_{h=1}^H\lVert\vx_{t+h}^i\rVert_2^2$.}
of the prediction residual
\begin{equation}
\vr_t^i \doteq \vx_{t:t+H}^i - \hat \vx_{t:t+H}^i \in\mathbb R^{H\times d_x}, 
\end{equation}
that is, the difference between the observed target data $\vx_{t:t+H}^i$ at the $i$-th node and the associated prediction $\hat \vx_{t:t+H}^i$ of model $\mathcal{F}$ made at time step $t-1$.
We stress that the residual $\vr_t^i$ should be interpreted as a $(Hd_x)$-dimensional vector, rather than a time series of length $H$, with each component associated with predictions made at time step $t-1$; for instance, considering one-step-ahead forecasts, $H=1$ and $\vr_t^i$ refers to $d_x$-dimensional forecast of $\vx_{t}^i$.
Given a test sequence, the mean squared error
\begin{equation}
\textrm{MSE}(\mathcal F)\doteq \frac{1}{T\,N}\sum_{t=1}^T\sum_{i=1}^N\lVert \vr_t^i\rVert_2^2
\end{equation}
is an estimate of $\mathbb E\left[\lVert\vr^i_{t}\rVert_2^2\right]$. 
Other common performance metrics are the \gls{rmse}~(computed as $ \sqrt{\textrm{MSE}(\mathcal F)}$), the \gls{mae} considering the 1-norm instead of the 2-norm, the \gls{mape} weighing each residual norm $\lVert \vr_t^i\rVert_1$ by the observed target $\lVert \vx_{t:t+H}^i\rVert_1$, and the \gls{mre} normalizing the sum absolute errors by sum of the observed values. Indeed, the best-performing model can be different depending on the considered figure of merit. 

Limiting the evaluation to computing a set of performance metrics suffers from one important drawback. Model performance does not provide any information about the possible room for improvement and does not allow for assessing whether or not the model can be considered optimal. 
This limitation emerges in all real-world scenarios, as the optimal achievable performance is unknown. A possible way out -- where graph-based processing can be pivotal, as discussed below -- consists in assessing the presence of correlations among residuals to complement the performance-based evaluation. The underlying principle is that correlated residuals indicate the presence of structural information not captured by the model \citep{li2003diagnostic}, thus suggesting that the predictions can be improved.
Several statistical hypothesis tests to detect the presence of residual dependencies have been conceived over the years~\citep{durbin1950testing, ljung1978measure, box2015time} and are mainly referred to as \emph{randomness} tests or \emph{whiteness} tests; these terms are reminiscent of white noise, i.e., a stochastic process displaying no correlations w.r.t.\ different points in time and space thus being completely random. 
Given residuals $\{\vr_t^i\}$, such tests typically compute a statistic $C(\{\vr_t^i\})$ in the form of a weighted sum of pairwise scalar statistics $\kappa(\vr_{t}^{i}, \vr_{s}^{j})$ between residuals $\vr_{t}^{i}, \vr_{s}^{j}$. Whenever the absolute value of test statistic $C(\{\vr_t^i\})$ is larger than a threshold $\gamma$, correlation is considered to be significant, that is
\begin{equation}
\text{If } \left|C\left(\left\{\vr_t^i\right\}\right)\right| > \gamma \implies \text{ reject hypothesis of uncorrelated residuals.}
\end{equation}
While whiteness tests do not quantify the margin for model improvement, they overcome the limits of performance-based analyses by providing a global assessment independently from specific performance metrics and baselines. The following discusses how spatiotemporal relational structure can help in performing such a correlation analysis in collections of correlated time series.

\paragraph{Testing for residual spatiotemporal correlations} Most previous works focused on analyzing correlations along the temporal dimension \citep{hosking1981equivalent,li2019testing,bose2020whiteness} or among different time series~(sensors)~\citep{moran1950notes,cliff1970spatial}. The joint analysis of spatiotemporal correlation however is more challenging, as studying the correlation between all possible residual pairs scales quadratically with both $N$ and $T$. 
In such a setting, the relational side information associated with the time series -- the graphs defined by $\{\gE_t\}$ -- has enabled spatiotemporal correlation analyses that scale to large time series collections. In particular, \citet{zambon2022aztest} introduce the AZ-whiteness test by designing the test statistic
\begin{align}\label{eq:az-test-stat}
    C(\{\vr_t^i\}) = \sum_{t}\sum_{(i,j) \in \gE_t} \alpha_{t}^{ij} \,\kappa(\vr_{t}^{i}, \vr_{t}^{j}) + \sum_{t}\sum_{i} \beta_{t}^{i}\, \kappa(\vr_{t}^{i}, \vr_{t+1}^{i}) 
\end{align}
composed of a sum over \emph{spatial} edges in $\{\gE_t\}$ and a sum over \emph{temporally} consecutive residuals. 
Scalars $\alpha_{t}^{ij}, \beta_{t}^{i}$ account for edge weights (e.g., the strength of the relations between time series) and trade off the weight given to spatial and temporal correlations, respectively.
The scalar statistic $\kappa({}\cdot{},{}\cdot{})$ is defined as
\begin{equation}\label{eq:az-test-local-stat}
    \kappa(\vr_{t}^{i}, \vr_{t}^{j})=\sign\left(\left\langle \vr_{t}^{i}, \vr_{s}^{j}\right\rangle\right)
\end{equation}
with $\sign({}\cdot{})$ being the sign function%
    \footnote{The sign function is such that $\sign(a)$ is equal to $-1$, $0$ or $1$ if $a$ is positive, null or negative, respectively.}
and $\langle{}\cdot{},{}\cdot{}\rangle$ the scalar product. 
Under mild assumptions, the test statistic in \autoref{eq:az-test-stat} has been proven to be asymptotically distributed as a standard Gaussian distribution~\citep{zambon2022aztest} so that $\gamma$ can be easily selected to meet desired confidence levels. 
The AZ-whiteness test has two main advantages. First, the test is scalable as it confines the correlation analysis only to residual pairs that are more likely to be correlated, i.e., those close in time or space~(e.g., $(\vr_{t}^{i},\vr_{t+1}^{i})$ and $(\vr_{t}^{i}, \vr_{t}^{j})$ with $(i,j)\in\gE_t$ as shown in \autoref{eq:az-test-stat}). As a result, the computation of test statistic $C(\{\vr_t^i\})$ scales linearly in the number of edges and time steps.
Second, thanks to the use of function $\sign({}\cdot{})$ in \autoref{eq:az-test-local-stat}, the test is distribution-free, which enables its application to real-world scenarios where the distribution of the residuals is typically unknown.

Besides global assessments of the overall model quality, the analysis of residual patterns localized in space and/or time provides valuable insights for discovering issues related to, e.g., faulty sensors, non-stationary dynamics, or dependencies that the model could not properly learn~\citep{cliff1970spatial}. \citet{anselin1995local} pioneered research in this direction, although focusing on spatial data for geographical analysis. \citet{zambon2023where}, instead, provide a set of procedures, based on the AZ-whiteness statistics, to identify space-time regions associated with significant residual correlations or inaccurate forecasts.  

\section{Practical examples and experiments}\label{sec:experiments}

Before analyzing challenges and future directions, this section complements the discussion carried out so far with numerical simulations on benchmark datasets from relevant application domains and synthetic data. The objective here is to show the impact of the transition from standard global and local deep learning predictors to graph-based architectures when forecasting collections of correlated time series. We follow the same experimental settings of \citet{cini2023taming}.

\paragraph{Baselines} As a case study, we consider \emph{recurrent} architectures. In particular, starting from standard \glspl{rnn}, implemented as \glspl{gru}~\citep{chung2014empirical}, we compare the performance of a single global \gls{rnn} sharing parameters across the collections against a set of local models and against the multivariate approach.
In particular, we consider the following baselines.
\begin{description}[leftmargin=1em, itemindent=1em]
    \item[\textbf{\acrshort{rnn}}:] a global node-level \gls{gru} conditioning predictions only on the history of the target as in \autoref{eq:global-node-level}. This model does not take spatial dependencies into account.
    \item[\textbf{\gls{fcrnn}}:] a \gls{gru} taking as input all of the time series concatenated along the spatial dimension as if they were a single multivariate sequence. This model lacks flexibility and does not exploit prior relational information.
    \item[\textbf{\gls{local_rnn}}:] a set of local \glspl{gru}. Each \gls{gru} is specialized on a specific time series and no parameter is shared. Similarly to the global node-level model, spatial dependencies are ignored. 
\end{description}
Then, for what concerns graph-based architectures, we consider both \gls{tts} and \gls{tas} recurrent architectures. Specifically, we build \gls{tts} models by stacking \gls{mp} layers after a \gls{rnn} encoder and take \glspl{gcrnn} as reference \gls{tas} architectures. For both architectures, we implement variants with both isotropic and anisotropic \acrlong{mp}. In particular, we compare the following model architectures.
\begin{description}[leftmargin=1em, itemindent=1em]
    \item[\textbf{\gls{ttsimp}}:] a global \gls{tts} model composed by a \gls{gru} followed by a stack of isotropic \gls{mp} layers. The \gls{mp} operator is defined as in \autoref{eq:imp}.
    \item[\textbf{\gls{ttsamp}}:] a global \gls{tts} model composed by a \gls{gru} followed by anisotropic \gls{mp} layers. The \gls{mp} operator is defined as in \autorefseq{eq:amp_first}{eq:amp_last}.
    \item[\textbf{\gls{tasimp}}:] a global \gls{tas} gated \gls{gcrnn} with isotropic \gls{mp}. The recurrent cell implementation follows~\autorefseq{eq:gcrnn-first}{eq:gcrnn-last}, the \gls{mp} operator is set up as in \autoref{eq:imp}.
    \item[\textbf{\gls{tasamp}}:] a global \gls{tas} gated \gls{gcrnn} with anisotropic \gls{mp}. The recurrent cell implementation follows~\autorefseq{eq:gcrnn-first}{eq:gcrnn-last}, the \gls{mp} operator is set up as in \autorefseq{eq:amp_first}{eq:amp_last}.
\end{description}
All the considered architectures follow the schema defined in \autorefseq{eq:enc}{eq:dec}, and the different variants are obtained by changing the implementation of the \gls{stmp} block. We stress again that all the global models share the same parameters across the time series in the collection. Finally, we also consider global-local variants of the above global models~(\gls{rnn} included) by adding node embeddings as inputs to the encoder and/or decoder, as in \autoref{eq:node-emb-enc} and \autoref{eq:node-emb-dec}. 

\subsection{Synthetic data}

In this experiment, we show the models' performance in a controlled environment. We train the models on the task of one-step-ahead prediction. We use the \gls{mae} as the figure of merit and report the AZ-whiteness statistics.

\paragraph{System model}
We consider the variation of GPVAR~\citep{zambon2022aztest} provided by \citet{cini2023taming} as the data-generating process. Data are generated by the recursive application, starting from noise, of a polynomial graph filter~\citep{isufi2019forecasting} (with parameters shared across time series) and an autoregressive filter (with parameters specific to each time series). Formally, the underlying system model is specified by
\begin{align}\label{eq:gpvar}
    \mH_t &= \sum_{l=1}^L\sum_{q=1}^{Q} \Theta_{q,l}\mA^{l-1}\mX_{t-q},\notag\\
    \mX_{t+1} &= \va \odot \text{tanh}\left(\mH_t\right) + \vb\odot\text{tanh}\left(\mX_{t-1}\right) + \eta_t,
\end{align}
where $\boldsymbol{\Theta}\in\sR^{Q \times L}$, $\va\in\sR^N$, $\vb\in\sR^N$ and $\eta_t \sim \mathcal{N}(\boldsymbol{0}, \sigma^2\sI)$. As in \citep{cini2023taming}, we consider two variants of the dataset, according to the initialization of $\va$ and $\vb$. In \textbf{\gls{lgpvar}} we set $\va$ and $\vb$ by sampling them from a uniform distribution as $\va,\vb \sim \mathcal{U}\left(-2, 2\right)$ to inject local effects into the process. In \textbf{\gls{gpvar}}, instead, we fix $\va=\vb=\boldsymbol{0.5}$ to remove any local effect. A detailed description of the experimental setting is reported in Appendix~\ref{a:datasets}.

% Note use of \abovespace and \belowspace to get reasonable spacing
% above and below tabular lines.
\begin{table}[t]
\caption{One-step-ahead forecasting error (\acrshort{mae}) of on GPVAR~(5 runs).}
\label{t:gpvar}
\small
\setlength{\tabcolsep}{1.2pt}
\setlength{\aboverulesep}{0pt}
\setlength{\belowrulesep}{0pt}
\renewcommand{\arraystretch}{1.25}
\begin{center}
% \resizebox{\textwidth}{!}{%
\begin{tabular}{c|c||r|r r r|r|r r r}
 \cmidrule[1pt]{3-10}
 \multicolumn{2}{c}{} & \multicolumn{4}{c|}{\textbf{\gls{gpvar}}} & \multicolumn{4}{c}{\textbf{\gls{lgpvar}}} \\
 \toprule
 \multicolumn{2}{c||}{\multirow{2}{*}{\textbf{MODELS}}} & \multicolumn{1}{c|}{\multirow{2}{*}{MAE}} & \multicolumn{3}{c|}{AZ-test} & \multicolumn{1}{c|}{\multirow{2}{*}{MAE}} & \multicolumn{3}{c}{AZ-test} \\
 % \cmidrule[.5pt]{4-6}\cmidrule[.5pt]{8-10}
 \multicolumn{2}{c||}{} & & \multicolumn{1}{c}{Time} & \multicolumn{1}{c}{T\texttt{+}S} & \multicolumn{1}{c|}{Space} & & \multicolumn{1}{c}{Time} & \multicolumn{1}{c}{T\texttt{+}S} & \multicolumn{1}{c}{Space} \\
\toprule
\multicolumn{2}{c||}{\acrshort{rnn}} & \cellcolor{gray!16}.3999{\tiny$\pm$.0000} & -3.0{\tiny$\pm$1.3} & \cellcolor{gray!50}35.7{\tiny$\pm$1.0} & \cellcolor{gray!50}53.5{\tiny$\pm$0.5} & \cellcolor{gray!46}.5441{\tiny$\pm$.0002} & \cellcolor{gray!16}10.8{\tiny$\pm$2.6} & 0.5{\tiny$\pm$1.9} & -10.1{\tiny$\pm$0.3} \\
\multicolumn{2}{r||}{\cellcolor{green!45!black!15} $\hookrightarrow$ \texttt{+} \footnotesize Emb.} & \cellcolor{gray!16}.3991{\tiny$\pm$.0000} & -2.6{\tiny$\pm$1.4} & \cellcolor{gray!50}34.7{\tiny$\pm$1.2} & \cellcolor{gray!50}51.7{\tiny$\pm$1.1}
 & \cellcolor{gray!29}.4611{\tiny$\pm$.0003} & \cellcolor{gray!6}6.1{\tiny$\pm$1.4} & -1.1{\tiny$\pm$1.1} & -7.7{\tiny$\pm$0.8} \\ \cmidrule{1-10}
\multicolumn{2}{c||}{\gls{fcrnn}} & \cellcolor{gray!24}.4388{\tiny$\pm$.0027} &\cellcolor{gray!50}261.0{\tiny$\pm$1.4} & \cellcolor{gray!50}252.2{\tiny$\pm$6.3} & \cellcolor{gray!50}95.6{\tiny$\pm$8.6}
 & \cellcolor{gray!50}.5948{\tiny$\pm$.0102} & \cellcolor{gray!50}108.4{\tiny$\pm$8.1} & \cellcolor{gray!50}73.6{\tiny$\pm$6.5} & -4.4{\tiny$\pm$2.3} \\ \cmidrule{1-10}
\multicolumn{2}{c||}{\gls{local_rnn}} & \cellcolor{gray!16}.4047{\tiny$\pm$.0001} & \cellcolor{gray!8}7.0{\tiny$\pm$3.7} & \cellcolor{gray!50}43.4{\tiny$\pm$4.2} & \cellcolor{gray!50}54.4{\tiny$\pm$2.3} & \cellcolor{gray!26}.4610{\tiny$\pm$.0003} & 3.2{\tiny$\pm$1.1} & -2.3{\tiny$\pm$1.1} & -6.5{\tiny$\pm$1.1} \\
\midrule
\multirow{4}{*}{\rotatebox[origin=c]{90}{\gls{tts}}} 
& \gls{ttsimp} & .3193{\tiny$\pm$.0000} & 0.9{\tiny$\pm$0.0} & 0.5{\tiny$\pm$0.7} & -0.3{\tiny$\pm$0.1} & \cellcolor{gray!12}.3808{\tiny$\pm$.0031} & \cellcolor{gray!22}13.8{\tiny$\pm$2.2} & \cellcolor{gray!10}7.9{\tiny$\pm$1.6} & -2.6{\tiny$\pm$0.9} \\
& \multicolumn{1}{r||}{\cellcolor{green!45!black!15} $\hookrightarrow$ \texttt{+} \footnotesize Emb.} & .3194{\tiny$\pm$.0000} & 2.8{\tiny$\pm$2.3} & 1.8{\tiny$\pm$1.7} & -0.2{\tiny$\pm$0.2} & .3197{\tiny$\pm$.0001} & 1.4{\tiny$\pm$1.0} & 1.0{\tiny$\pm$0.9} & -0.0{\tiny$\pm$0.3} \\ \cmidrule{2-10}
& \gls{ttsamp} & .3193{\tiny$\pm$.0000} & 1.2{\tiny$\pm$1.6} & 0.8{\tiny$\pm$1.1} & -0.1{\tiny$\pm$0.1} & \cellcolor{gray!9}.3639{\tiny$\pm$.0032} & \cellcolor{gray!21}13.1{\tiny$\pm$2.6} & \cellcolor{gray!9}7.5{\tiny$\pm$2.4} & -2.5{\tiny$\pm$1.0}\\
& \multicolumn{1}{r||}{\cellcolor{green!45!black!15} $\hookrightarrow$ \texttt{+} \footnotesize Emb.} & .3194{\tiny$\pm$.0000} & 1.4{\tiny$\pm$3.6} & 0.8{\tiny$\pm$2.5} & -0.2{\tiny$\pm$0.1} & .3199{\tiny$\pm$.0001} & 1.8{\tiny$\pm$0.7} & 1.0{\tiny$\pm$0.6} & -0.3{\tiny$\pm$0.3} \\
\midrule
\multirow{4}{*}{\rotatebox[origin=c]{90}{\gls{tas}}} 
& \gls{tasimp} & .3194{\tiny$\pm$.0000} & 1.9{\tiny$\pm$0.4} & 1.2{\tiny$\pm$0.4} & -0.3{\tiny$\pm$0.2} & \cellcolor{gray!10}.3714{\tiny$\pm$.0070} & \cellcolor{gray!25}15.2{\tiny$\pm$2.9} & \cellcolor{gray!12}9.0{\tiny$\pm$1.6} & -2.5{\tiny$\pm$1.5} \\
& \multicolumn{1}{r||}{\cellcolor{green!45!black!15} $\hookrightarrow$ \texttt{+} \footnotesize Emb.} & .3196{\tiny$\pm$.0000} & 0.8{\tiny$\pm$3.0} & 0.4{\tiny$\pm$2.1} & -0.3{\tiny$\pm$0.2} & .3204{\tiny$\pm$.0001} & 2.4{\tiny$\pm$0.9} & 1.8{\tiny$\pm$0.7} & 0.1{\tiny$\pm$0.2} \\ \cmidrule{2-10}
& \gls{tasamp} & .3195{\tiny$\pm$.0000} & 2.6{\tiny$\pm$2.0} & 1.7{\tiny$\pm$1.4} & -0.3{\tiny$\pm$0.2} & \cellcolor{gray!6}.3518{\tiny$\pm$.0013} & \cellcolor{gray!15}10.5{\tiny$\pm$2.5} & \cellcolor{gray!5}5.7{\tiny$\pm$1.9} & -2.4{\tiny$\pm$0.6} \\
& \multicolumn{1}{r||}{\cellcolor{green!45!black!15} $\hookrightarrow$ \texttt{+} \footnotesize Emb.} & .3197{\tiny$\pm$.0000} & 1.7{\tiny$\pm$2.6} & 1.2{\tiny$\pm$1.9} & -0.0{\tiny$\pm$0.2} & .3204{\tiny$\pm$.0002} & 1.8{\tiny$\pm$0.6} & 0.9{\tiny$\pm$0.4} & -0.4{\tiny$\pm$0.5} \\
\midrule
\multicolumn{2}{c||}{Optimal model} & .3192\hspace{.8cm} & ---\hspace{.25cm} & ---\hspace{.25cm} & ---\hspace{.25cm} & .3192\hspace{.8cm} & ---\hspace{.25cm} & ---\hspace{.25cm} & ---\hspace{.25cm} \\
\bottomrule
\end{tabular}%
% }
% \vspace{-.5cm}
\end{center}
\end{table} 

\paragraph{Results}
\autoref{t:gpvar} reports the models' forecasting performance in terms of \gls{mae} and three values of the AZ-whiteness test statistic~\citep{zambon2022aztest} accounting for temporal, spatial and spatiotemporal correlations%
\footnote{Statistics for temporal (spatial) correlations are obtained by setting all weights $\beta_t^i$ ($\alpha_t^{ij}$) to $0$.}. 
The performance of the optimal model is obtained analytically by considering the variance of the noise $\eta_t$ in \autoref{eq:gpvar}. As expected, models that do not exploit spatial dependencies~(\gls{rnn}, \gls{fcrnn} and \gls{local_rnn}) struggle in both datasets, displaying large residual spatial correlation, as shown by the spatial and spatiotemporal statistics.
Note that spatial correlations are more difficult to detect in \gls{lgpvar}, due to the presence of the local dynamics determined by random vectors $\va,\vb$ in \autoref{eq:gpvar}, which is reflected in the values of the spatiotemporal statistic as well, as it balances the temporal and spatial components.
Graph-based methods, instead, achieve performance close to the theoretical optimum in \gls{gpvar}, with the test statistics close to zero. 
For what concerns \gls{lgpvar}, global models~(including, the graph-based methods) struggle to account for the introduced local effects. Conversely, global-local graph-based methods achieve good results in both benchmarks. 

\subsection{Benchmarks}

This sequel of the section provides an assessment of the introduced baselines on datasets coming from real-world applications to show the performance of the discussed methodologies.

% Note use of \abovespace and \belowspace to get reasonable spacing
% above and below tabular lines.
\begin{table}[t]
\caption{Forecasting results on $4$ benchmark datasets (5 runs). Best model performance within each group~(local, global, global-local) reported in \textbf{bold}.}\label{t:benchmarks}
% \vspace{-.15cm}
\small
\setlength{\tabcolsep}{2pt}
\setlength{\aboverulesep}{0pt}
\setlength{\belowrulesep}{0pt}
\renewcommand{\arraystretch}{1.2}
\begin{center}
% \resizebox{\textwidth}{!}{%
\begin{tabular}{c|r r|r r|r r|r r|r r}
\cmidrule[1pt]{2-11}
 \multicolumn{1}{c}{} & \multicolumn{2}{c|}{\bfseries \gls{la}} & \multicolumn{2}{c|}{\bfseries \gls{bay}} & \multicolumn{2}{c|}{\bfseries \gls{cer}} & \multicolumn{2}{c|}{\bfseries \gls{air}} & \multicolumn{2}{c}{\bfseries \gls{engrad}} \\
\cmidrule[.5pt]{2-11}
\multicolumn{1}{c}{} & \multicolumn{1}{c}{MAE} & \multicolumn{1}{c}{MRE} & \multicolumn{1}{c}{MAE} & \multicolumn{1}{c}{MRE} & \multicolumn{1}{c}{MAE} & \multicolumn{1}{c}{MRE} & \multicolumn{1}{c}{MAE} & \multicolumn{1}{c}{MRE} & \multicolumn{1}{c}{MAE} & \multicolumn{1}{c}{MRE} \\ \toprule
% \multicolumn{4}{c|}{Global models}  & \multicolumn{4}{c}{Global-local models (with embeddings)} \\
\multicolumn{1}{c}{} & \multicolumn{10}{c}{\sc Local models} \\ \midrule
FC-RNN & \bftab{3.56{\tiny$\pm$.03}} & \bftab{6.16{\tiny$\pm$.04}} & 2.32{\tiny$\pm$.01} & 3.72{\tiny$\pm$.02} & 713.01{\tiny$\pm$8.27} & 33.75{\tiny$\pm$.39} & 18.24{\tiny$\pm$.07} & 28.45{\tiny$\pm$.11} & \bftab{55.37{{\tiny $\pm$1.57}}} & \bftab{23.15{{\tiny $\pm$.66}}} \\
LocalRNNs & 3.69{\tiny$\pm$.00} & 6.38{\tiny$\pm$.01} & \bftab{1.91{\tiny$\pm$.00}} & \bftab{3.06{\tiny$\pm$.00}} & \bftab{508.95{\tiny$\pm$1.48}} & \bftab{24.09{\tiny$\pm$.07}} & \bftab{14.75{\tiny$\pm$.02}} & \bftab{23.02{\tiny$\pm$.03}} & 58.80{{\tiny $\pm$0.31}} & 24.59{{\tiny $\pm$.13}} \\ \midrule
% \midrule
\multicolumn{1}{c}{} & \multicolumn{10}{c}{\sc Global models} \\ \midrule
RNN & 3.54{\tiny$\pm$.00} & 6.13{\tiny$\pm$.00} & 1.77{\tiny$\pm$.00} & 2.84{\tiny$\pm$.00} & 456.98{\tiny$\pm$0.61} & 21.63{\tiny$\pm$.03} & 14.02{\tiny$\pm$.04} & 21.87{\tiny$\pm$.07} & 47.41{{\tiny $\pm$0.57}} & 19.82{{\tiny $\pm$.24}} \\
\midrule
\gls{ttsimp} & 3.34{\tiny$\pm$.01} & 5.79{\tiny$\pm$.01} & 1.72{\tiny$\pm$.00} & 2.75{\tiny$\pm$.01} & 439.13{\tiny$\pm$0.51} & 20.79{\tiny$\pm$.02} & 12.74{\tiny$\pm$.02} & 19.88{\tiny$\pm$.04} & \bftab{44.48{{\tiny $\pm$0.24}}} & \bftab{18.60{{\tiny $\pm$.10}}} \\
\gls{ttsamp} & \bftab{3.24{\tiny$\pm$.01}} & \bftab{5.61{\tiny$\pm$.01}} & 1.66{\tiny$\pm$.00} & \bftab{2.65{\tiny$\pm$.01}} & \bftab{431.33{\tiny$\pm$0.68}} & \bftab{20.42{\tiny$\pm$.03}} & \bftab{12.30{\tiny$\pm$.02}} & \bftab{19.20{\tiny$\pm$.03}} & \bftab{44.62{{\tiny $\pm$0.35}}} & \bftab{18.66{{\tiny $\pm$.15}}} \\
\gls{tasimp} & 3.35{\tiny$\pm$.01} & 5.80{\tiny$\pm$.01} & 1.70{\tiny$\pm$.01} & 2.73{\tiny$\pm$.01} & 443.85{\tiny$\pm$0.99} & 21.01{\tiny$\pm$.05} & 12.87{\tiny$\pm$.02} & 20.08{\tiny$\pm$.04} & 45.55{{\tiny $\pm$0.33}} & 19.05{{\tiny $\pm$.14}} \\
\gls{tasamp} & \bftab{3.22{\tiny$\pm$.02}} & \bftab{5.58{\tiny$\pm$.03}} & \bftab{1.65{\tiny$\pm$.00}} & \bftab{2.64{\tiny$\pm$.00}} & 456.72{\tiny$\pm$3.91} & 21.62{\tiny$\pm$.19} & \bftab{12.29{\tiny$\pm$.02}} & \bftab{19.18{\tiny$\pm$.04}} & \bftab{43.93{{\tiny $\pm$0.55}}} & \bftab{18.37{{\tiny $\pm$0.23}}} \\
\midrule
\multicolumn{1}{c}{} & \multicolumn{10}{c}{\sc Global-local models (with embeddings)} \\ \midrule
RNN & 3.15{\tiny$\pm$.03} & 5.45{\tiny$\pm$.05} & 1.59{\tiny$\pm$.00} & 2.54{\tiny$\pm$.00} & 421.50{\tiny$\pm$1.78} & 19.95{\tiny$\pm$.08} & 13.73{\tiny$\pm$.04} & 21.42{\tiny$\pm$.06} & 46.83{{\tiny $\pm$0.19}} & 19.58{{\tiny $\pm$.08}} \\
\midrule
\gls{ttsimp} & \bftab{3.08{\tiny$\pm$.01}} & \bftab{5.33{\tiny$\pm$.03}} & \bftab{1.58{\tiny$\pm$.00}} & \bftab{2.53{\tiny$\pm$.00}} & \bftab{412.44{\tiny$\pm$2.80}} & \bftab{19.52{\tiny$\pm$.13}} & 12.33{\tiny$\pm$.02} & 19.24{\tiny$\pm$.03} & 43.96{{\tiny $\pm$0.42}} & 18.38{{\tiny $\pm$.17}} \\
\gls{ttsamp} & \bftab{3.06{\tiny$\pm$.01}} & \bftab{5.29{\tiny$\pm$.02}} & \bftab{1.58{\tiny$\pm$.01}} & \bftab{2.54{\tiny$\pm$.01}} & \bftab{412.95{\tiny$\pm$1.28}} & \bftab{19.55{\tiny$\pm$.06}} & \bftab{12.15{\tiny$\pm$.02}} & \bftab{18.96{\tiny$\pm$.03}} & 43.70{{\tiny $\pm$0.33}} & 18.27{{\tiny $\pm$.14}} \\
\gls{tasimp} & 3.10{\tiny$\pm$.01} & 5.36{\tiny$\pm$.02} & 1.59{\tiny$\pm$.00} & 2.55{\tiny$\pm$.00} & 417.71{\tiny$\pm$1.28} & 19.77{\tiny$\pm$.06} & 12.48{\tiny$\pm$.03} & 19.47{\tiny$\pm$.04} & 44.90{{\tiny $\pm$0.33}} & 18.77{{\tiny $\pm$.14}} \\
\gls{tasamp} & \bftab{3.07{\tiny$\pm$.02}} & \bftab{5.31{\tiny$\pm$.04}} & 1.59{\tiny$\pm$.00} & \bftab{2.54{\tiny$\pm$.01}} & 416.74{\tiny$\pm$1.57} & 19.73{\tiny$\pm$.07} & \bftab{12.17{\tiny$\pm$.05}} & \bftab{18.98{\tiny$\pm$.08}} & \bftab{43.14{{\tiny $\pm$0.19}}} & \bftab{18.04{{\tiny $\pm$.08}}} \\
\bottomrule
\end{tabular}%
% } 
\end{center}
\end{table}

\paragraph{Datasets} Following \citet{cini2023taming}, we consider benchmarks coming from traffic forecasting, energy analytics and air quality monitoring domain. In particular, we use the following datasets.

\begin{description}[leftmargin=1em, itemindent=1em]
    \item[\gls{la} \& \gls{bay}] \Gls{la} and \gls{bay}, introduced by ~\citet{li2018diffusion}, are two popular traffic forecasting datasets consisting of measurements from loop detectors in the Los Angeles County Highway and San Francisco Bay Area, respectively~\citep{chen2001freeway}.
    \item[\textbf{\gls{cer}}] The \gls{cer} dataset~\citep{cer2016cer} consists of energy consumption readings, aggregated into $30$-minutes intervals, from $485$ smart meters monitoring small and medium-sized enterprises.
    \item[\textbf{\gls{air}}] The \gls{air} dataset~\citep{zheng2015forecasting} collects hourly measurements of pollutant PM2.5 from $437$ air quality monitoring stations in China, spread across different cities. 
    \item[\textbf{\gls{engrad}}] Introduced by~\citet{marisca2024graph}, the \gls{engrad} dataset contains three years of $5$ hourly-sampled weather variables generated by ECMWF Integrated Forecasting System~(IFS). %\footnote{\url{https://www.ecmwf.int/}}
    for $487$ grid points in England. For our experiment, we use shortwave radiation as the target variable $\vx^i_t$ and the remaining four variables as exogenous $\vu^i_t$.
\end{description}
We use the same data splits and preprocessing of previous works~\citep{wu2019graph, cini2023taming}. In particular, the adjacency matrices for the traffic and air quality monitoring datasets are obtained by applying a thresholded Gaussian kernel~\citep{shuman2013emerging} on the pairwise geographic distances among sensors; for \gls{cer}, following previous works~\citep{cini2022filling}, the graph connectivity is derived from the correntropy~\citep{liu2007correntropy} among time series. We refer to Appendix~\ref{a:datasets} for more details.

\paragraph{Results} \autoref{t:benchmarks} shows the results of the empirical evaluation of the reference models on the selected datasets. Graph-based architectures outperform standard local and global predictors in all considered scenarios; the performance gap is particularly wide when considering fully global models. As one might expect, local models perform and scale poorly. This is particularly evident in \gls{engrad}, where data are generated from similar processes. In this scenario, local components bring limited benefits compared to a fully global architecture. In every other case, hybrid global-local models obtain markedly better performance than fully global baselines. However, it should be noted that such models lack flexibility in inductive settings as discussed in \autoref{sec:inductive}. Moreover, anisotropic message-passing schemes outperform their isotropic counterparts in most scenarios, while \gls{tts} architectures perform on par or better than \gls{tas} models. 
Finally note that, although the above results are significant, they do not necessarily generalize to all datasets and \gls{tts}/\gls{tas} architectures.

\section{Challenges}\label{sec:challenges}

This section identifies and discusses the main challenges that the practitioner would typically have to deal with in processing collections of time series with graph-based forecasting methods.

\subsection{Dealing with missing data}\label{sec:imputation}

The time series in the collection, i.e., $\mX_{t:t+T}$, may be affected by missing values, as pointed out in \autoref{sec:extension}. Phenomena that result in incomplete observations include~(among others) irregular sampling procedures, acquisition and communication errors, and hardware and software faults. Moreover, it is often the case that the time frames of the time series in the collection do not overlap perfectly, e.g., sensors might be installed at different points in time. This section provides guidelines on dealing with incomplete observations in settings where dependencies among the time series in the collection can be exploited for reconstruction. In particular, we discuss how the graph-based methodologies presented in the paper provide useful tools to tackle the problem. For a complete treatment of these aspects, we refer to~\citet{cini2022filling} and \citet{marisca2022learning}.

\paragraph{Graph Deep Learning for Time Series Imputation} Although incomplete, we assume all the available time series to be synchronous and regularly sampled and consider the masked representation introduced in \autoref{sec:extension}. In particular, we pair each $\gG_t$ with a binary mask $\mM_t$ to indicate the missing observations. To simplify the presentation, we do not consider partial observability at the level of the single sensor, i.e., given an observation vector $\vx^i_t$, either all the channels are observed or none is available, i.e., $\vm^i_t \in \{\vzero, \vone \}$. However, no further assumption is made about the missing data distribution. 
Clearly, the gaps in the observed data must be accounted for while processing the data.
A common approach consists of reconstructing missing observations before carrying out the downstream processing, by exploiting some imputation model. Besides standard statistical methods, deep learning approaches have become a popular alternative~\citep{lipton2016directly, yoon2018gain, cao2018brits}. In particular, graph deep learning offers the tools to exploit dependencies among time series in this context as well~\citep{cini2022filling, marisca2022learning, omidshafiei2022multiagent, chen2022adaptive, liu2023pristi, wang2023networked}. Indeed, \glspl{stgnn} have been successfully applied to multivariate time series imputation in the presence of relational side information, with attention-based methods gaining traction by solving error-compounding issues typical of recurrent architectures~\citep{marisca2022learning}. 
\citet{cini2022filling} formalize the reconstruction problem in the context of graph-based representations and provide a bidirectional \gls{gcrnn} -- paired with an additional spatial decoder -- that reconstructs the missing observations by exploiting both spatial and temporal dependencies. Indeed, the spatial decoder designed in \citep{cini2022filling} offers an example of how relational inductive biases can be exploited for data reconstruction. In particular, representations w.r.t.\ each $\vx^i_t$ vector can be obtained through \gls{stmp} by masking out unavailable past observations; representations can then be used for reconstruction by aggregating values observed at neighboring nodes, i.e., as
\begin{align}
    \mZ_{t}^{\ } &= \textsc{STMP}\left(\mH^{\ }_{<t} \odot \mM^{\ }_{<t},\gE^{\ }_{<t} \right),\\ 
    \widehat\vx^{i}_{t} &= \textsc{Dec}\left(\vz_t^i, \aggr_{j \in \gN(i)\setminus \{i\}}\left\{ \msg(\vz^j_t, \vx^j_t \odot \vm^j_t) \right\}\right),
\end{align} 
where $\widehat\vx^{i}_{t}$ denotes the reconstructed signal and $\textsc{Decoder}$ a generic readout layer. The reconstruction can be conditioned on both past and future values by exploiting, e.g., a bidirectional architecture. Clearly, many possible designs are possible and research on the topic is increasingly active~(see \citep{jin2023survey}).

\paragraph{Forecasting from Partial Observations} A different and more direct approach to the problem is to avoid the reconstruction step and to consider forecasting architecture that can directly deal with irregular observations. Although research on the topic of graph-based methods in this context is limited~\citep{zhong2021heterogeneous}, many of the mechanisms used in imputation models to process the incomplete observations can potentially be adapted to build forecasting architectures~(see, e.g., \cite{marisca2022learning}). The main advantage of such adaptations is that the resulting model, trained end-to-end, could be used to jointly impute missing observations and forecast future values. Other methods, instead, tackle this problem in the context of continuous-time modeling, we discuss them in \autoref{sec:future}.

\subsection{Latent Graph Learning}\label{sec:graph-learning}

\Glspl{stgnn} rely on propagating representations through the spatial connections encoded in the graph that comes with the time series collection. The available relational information, however, can be inaccurate or inadequate for modeling the relevant dependencies. For instance, in neurobiology, the physical proximity of brain regions does not always explain the observed dynamics~\citep{van2014functional, grattarola2022seizure}. In other cases, relational information might be completely missing. Nonetheless, relational architectural biases can be exploited by learning a graph end-to-end with the forecasting architecture~\citep{kipf2018neural, wu2019graph, shang2021discrete, wu2020connecting, cini2023sparse}; in some sense, graph learning can be seen as a regularization of attention-based architectures~\citep{vaswani2017attention}, where, rather than relying on attention scores between each pair of nodes, the learned graph is used to route information only between certain nodes, thus providing localized node representations typical of graph-based processing.
Learning discrete representations~\citep{niculae2023discrete} while keeping computations sparse is indeed a key challenge for graph learning, with a large impact on the scalability of the resulting forecasting architecture~\citep{cini2023sparse}. The following paragraphs provide a critical overview of the most common approaches. 

\paragraph{Directly learning an adjacency matrix} Most \glspl{stgnn} rely on learning an adjacency matrix $\widetilde \mA$ as a function of a matrix of edge scores $\mPhi \in \sR^{N\times N}$ as 
\begin{equation}\label{eq:graph-learning-scores}
    \widetilde \mA = \xi\left(\mPhi\right) \quad \text{with learnable parameters} \quad \mPhi, 
\end{equation}
where $\xi({}\cdot{})$ indicates a nonlinear activation function that can be used to induce sparsity in the resulting adjacency matrix $\widetilde \mA$, e.g., by thresholding the scores s.t. $[\mA]_{ij}=1$ if $\phi_{ij}>\varepsilon$ and $0$ otherwise.
The cost of parametrizing the full score matrix $\mPhi$ can be amortized by factorizing it as
\begin{align}\label{eq:factorized-adj}
    \widetilde \mA = \xi\left(\mPhi\right) \quad \text{with} \quad \mPhi = \mZ_{src}\mZ_{tgt}^\top,
\end{align}
where $\mZ_{src},\mZ_{tgt} \in \sR^{N\times d_z}$ are node embeddings obtained, e.g., as a function of the available data or as tables of learnable parameters. Such factorization approach has been pioneered in the context of \glspl{stgnn} by the Graph WaveNet architecture~\citep{wu2019graph}, where $\xi({}\cdot{})$ is implemented by a ReLU followed by a row-wise softmax activation and node embeddings are learnable parameters. Several other methods have followed this direction~\citep{bai2020adaptive, oreshkin2020fcgaga, liu2022multivariate} which is quite flexible; indeed, making the embeddings dependent on the input window can easily allow for modeling dynamic relationships~\citep{kipf2018neural}, e.g., as 
  \begin{align}\label{eq:phi}
     \widetilde \mA_t = \xi\left(\mPhi_t\right)\quad\text{with}&\quad\mPhi_t = \left(\mZ_t \mW_{src}\right) (\mZ_t \mW_{tgt})^\top,\\ \vz^i_t = \textsc{SeqEnc}&\left(\vx^i_{t-W: t}, \vu^i_{t-W: t},
     \vv^i\right),
 \end{align}
 where $\textsc{SeqEnc}(\cdot)$ indicates a generic sequence encoder (e.g., an \gls{rnn}) and ${\mW_{src},\mW_{tgt}\in \sR^{d_z\times d}}$ are learnable weight matrices. The drawback of such methods is that they often result in a dense $\widetilde \mA$ matrix which makes any subsequent \gls{mp} operation scale with $\bigO(N^2)$ rather than with $\bigO(|\gE|)$. MTGNN~\citep{wu2020connecting} and GDN~\citep{deng2021graph} sparsify the learned factorized adjacency by selecting, for each node, the $K$ edges associated with the largest weights, which, however, results in sparse gradients. More recently, \citet{zhang2022graph} proposed a different approach based on the idea of sparsifying the learned graph by thresholding the average of learned attention scores. Finally, a general approach to learning dynamic edge scores is to compute them directly as a function of source and target node representations $\vz^i_{t}$ and $\vz^j_{t}$~\citep{cini2023sparse}, e.g,
\begin{equation}
    \widetilde \mA_t = \xi\left(\mPhi_t\right)\quad\text{with}\quad\mPhi_t[i,j] = \mlp(\vz^i_{t}, \vz^j_{t}).
\end{equation}

\paragraph{Learning graph distributions} A different approach to the graph learning problem consists of adopting a probabilistic perspective. Instead of directly learning a graph, probabilistic methods learn a probability distribution over graphs $p_\mPhi(\mA)$ such that graphs sampled from $p_\mPhi$ maximize the performance at task. The probabilistic approach allows for the embedding of priors directly into $p_\Phi$, enabling the learning of sparse graphs as realizations of a discrete probability distribution. For example, one can consider graph distributions $p_{\mPhi}$ such that
\begin{align}\label{eq:prob-graph-learning}
    \widetilde \mA_t \sim p_{\mPhi_t}(\mA_t),
\end{align}
where $p_{\mPhi_t}$ is parameterized by edge scores $\mPhi_t$ obtained adopting any of the parameterizations discussed in the previous paragraph. The graph distribution can be, e.g., implemented by considering a Bernoulli variable associated with each edge or by considering more complex distributions such as top-$K$ samplers~(see, e.g., \citep{cini2023sparse, kazi2022differentiable, paulus2020gradient, ahmed2023simple}).
Among probabilistic methods, NRI~\citep{kipf2018neural} introduces a latent variable model for predicting the interactions of physical objects by learning the discrete and dynamic edge attributes of a fully connected graph. GTS~\citep{shang2021discrete} simplifies the NRI module by considering only binary relationships and integrates the graph inference module in a recurrent \gls{stgnn}~\citep{li2018diffusion}. 
To learn $p_\Phi$, Both NRI and GTS exploit path-wise gradient estimators based on the categorical \emph{Gumbel trick}~\citep{maddison2017concrete, jang2017categorical}; as such, they rely on continuous relaxations of discrete distributions and suffer from the same computational setbacks of previously discussed methods. Recently, \citet{cini2023sparse} propose variance-reduced score-based estimators that allow for sparse \gls{mp} operations with $\bigO(|\gE|)$ computational complexity. 

Outside of applications to time series forecasting, \citet{franceschi2019learning} propose a bi-level optimization routine to learn graphs based on a straight-through estimator~\citep{bengio2013estimating}. \citet{kazi2022differentiable} uses the Gumbel-Top-K trick~\citep{kool2019stochastic} to sample a $K$-nearest neighbors ($K$-NN) graph and learn edge scores by using a heuristic for increasing the likelihood of sampling edges contributing to correct classifications. \citet{wren2022learning} learn DAGs end-to-end by exploiting implicit maximum likelihood estimation~\citep{niepert2021implicit}. In summary, the graph learning problem is inherently complex due to challenges related to both computational complexity as well as the learning of discrete representations with neural networks. We refer to  \citet{niculae2023discrete} and \citet{mohamed2020monte} for an in-depth discussion on methods and estimators for learning latent (discrete) structures in machine learning. 

\subsection{Computational Scalability}\label{sec:scalability}

In the problems considered so far, scalability concerns can emerge from both the number of time series in the collection as well as their length. 
Indeed, data span both the temporal and the spatial dimensions. In real-world applications, e.g., smart transportation systems in large cities, dealing with thousands of time series acquired at high sampling rates over long periods of time is rather common~\citep{cini2023scalable, liu2023largest}. This results in a large amount of data that needs to be processed at once to account for long-range spatiotemporal dependencies across the time series in the collection.
When designing and/or implementing an \gls{stgnn}, the scalability issue, then, must be taken into account. As mentioned in \autoref{sec:stgnn-architectures}, a generic \gls{tas} model performs $L$ stacked \gls{mp} operations for each time step resulting in a time and space complexity scaling with $\bigO\big(W(N + L|\gE_{\max}|)\big)$, or $\bigO\big(WL|\gE_{\max}|\big)$ assuming $N \ll |\gE_{\max}|$. \gls{stt} models are characterized by an analogous computation complexity, as the decoupled processing generally does not bring any advantage in this direction. Conversely, \gls{tts} models, by encoding the time series ahead of any \gls{mp} operation, scale with $\bigO\big(WN + L|\gE_t|\big)$, which, again assuming $N \ll |\gE_t|$, is a notable improvement. However, even models following this paradigm can struggle whenever either $N$, $W$, or $|\gE|$ are large, and appropriate computational resources can quickly become unaffordable. This issue is particularly relevant at training time when processing batches of such high-dimensional data concurrently is needed to fit \glspl{stgnn}' parameters on the available data. In the following, we discuss available methods to scale existing approaches to extremely large sensor networks, highlighting the shortcomings and advantages of the different approaches. 

\paragraph{Graph subsampling and clustering} An often viable solution is to subsample the data fed to the model. 
In particular, the computational burden can be reduced at training time by extracting subgraphs from the full-time series collection~\citep{hamilton2017inductive, chiang2019cluster, zeng2020graphsaint} by, e.g., considering the $K$-th order neighborhood of a subset of nodes. Such approaches have been exploited, mostly adapted from methods developed in the context of static graph processing, and have indeed been applied to scale graph-based time series forecasting to large networks~\citep{wu2020connecting, gandhi2021spatio, rong2020dropedge}. Subsampling methods, then, allow for capping the number of nodes/edges to be processed for each sample based on the available computational resources. The drawback of these approaches is that such a subsampling might break long-range spatiotemporal dependencies~(n.b., data are not i.i.d.) and result in a noisy learning signal~\citep{cini2023scalable}, i.e., high-variance gradient estimates. Similar arguments can be made w.r.t.\ small batch sizes and short input windows. As an alternative, other approaches reduce the computational complexity of processing the full graph by relying on graph clustering and pooling~\citep{grattarola2022understanding, bianchi2023expressive} to operate on hierarchical representations of the graph~\citep{yu2019st, cini2023graph}, but still require to process the full graph at the input and output layers.

\paragraph{Precomputing spatiotemporal encodings} Finally, a successful and popular approach to scale \glspl{gnn} to large graphs is to precompute a representation for each node ahead of training and then process the data as if they were i.i.d.~(e.g., see~\citep{frasca2020sign}). Such an approach has been extended to spatiotemporal data in~\citep{cini2023scalable} by exploiting randomized deep echo state networks~\citep{gallicchio2017deep, bianchi2020reservoir} and powers of a graph shift operator to extract, in an unsupervised fashion, spatiotemporal representations w.r.t. each time step and node before performing any training. The obtained representations can then be sampled uniformly across both time and space to efficiently train a decoder for mapping them to predictions~\citep{cini2023scalable}. The advantage of the preprocessing approach is that it makes the computational cost of a training step independent from both the length of the sequence and the number of nodes and edges by delegating the propagation of representation through time and space to the training-free encoding step. This encoding can be carried out only once before any training epoch, without being limited, e.g., by GPU memory availability. Clearly, although empirical performance matches the state of the art in relevant benchmarks~\citep{cini2023scalable}, the downside is that separating encoding and decoding can be less effective in certain scenarios and more reliant on hyperparameter selection compared to end-to-end approaches.

\subsection{Inductive Learning}\label{sec:inductive} 

As previously mentioned, global \glspl{stgnn} can make predictions for never-seen-before node sets, and handle graphs of different sizes and variable topology. In practice, graph-based predictors can be used for zero-shot transfer and inductive learning and can easily handle new time series being added to the collection which, for example, corresponds to the real-world scenario of new sensors being added to a network over time. The flexibility of these models has several applications in time series processing besides forecasting, e.g., as models for performing spatiotemporal kriging~\citep{stein1999interpolation} or virtual sensing~\citep{cini2022filling, wu2021inductive, zheng2023increase}, where inductive \glspl{stgnn} can be used to perform graph-based spatial interpolation. 
However, performance in the inductive setting can quickly degrade as soon as the target time series exhibit dynamics that deviate from those observed in the training examples~\citep{cini2023taming}. Furthermore, including node-specific local components in the forecasting architecture -- which as we discussed can be critical for accurate predictions -- makes such \glspl{stgnn} unable to perform inductive inferences. Luckily, as discussed in the following, adapting such models by exploiting a small number of observations can enable transfer.

\paragraph{Transfer learning} \glspl{stgnn} can be adjusted to account for other sets of time series~(with different dynamics) by fine-tuning on the available data a subset of the forecasting architectures weights~\citep{cini2023taming} or exploiting other transfer learning strategies~\citep{mallick2021transfer}, e.g., based on ideas from meta learning~\citep{panagopoulos2021transfer}. For what concerns global-local architectures, the use of node embeddings can amortize the cost of the transfer learning by limiting the fine-tuning of the model to fitting a new set of embeddings for the nodes in the target set while freezing the shared weights~\citep{cini2023taming}. Furthermore, node embeddings can be regularized to facilitate transfer by structuring the latent space~\citep{cini2023taming} or by forcing new node embeddings to be close to those learned from the initial training set~\citep{yin2022nodetrans}.

\section{Future Directions}\label{sec:future}

Besides the challenges identified in the paper that are indeed still open and the subject of extensive research, we can identify several promising research directions for future works to delve into.

\paragraph{Graph state-space models} Graph-based processing has been recently exploited to design state-space models~\citep{durbin2012time, rangapuram2018deep} based on a graph-structured state representations~\citep{zambon2023graph}. The resulting \textit{graph state-space models} learn state graphs that can be disjoint from the input, i.e., can have a number of nodes that is larger or smaller than the number of input time series and a different associated topology. Ad-hoc Kalman filtering techniques have been introduced and have led to promising empirical results~\citep{alippi2023graph}. The resulting framework encompasses several existing architectures~(e.g.,~\citep{seo2018structured, bai2020adaptive}) while enabling more advanced designs that, however, have not been fully explored yet. 

\paragraph{Spatial and temporal hierarchies} The spatiotemporal structure of the data allows for processing observations and making predictions at different scales, both in time~\citep{ athanasopoulos2017forecasting} and in space~\citep{hyndman2011optimal}. This idea has been indeed exploited in deep learning methods for hierarchical time series forecasting~\citep{rangapuram2021end, rangapuram2023coherent, challu2023nhits, zhou2023sloth, han2021simultaneously}. As briefly mentioned in \autoref{sec:scalability}, several \gls{stgnn} can take advantage of graph pooling to operate at different levels of resolution~\citep{yu2019st, guo2021hierarchical}. In particular, hierarchical time series forecasting and end-to-end graph-based time series clustering have been recently integrated within the same forecasting framework~\citep{cini2023graph}. \citet{marisca2024graph} used hierarchal representation as a means to deal with missing data and sparse observations. Future works can investigate methodologies to process spatiotemporal time series in an integrated hierarchical fashion across both time and space while taking the coherency of the forecasts into account. 

\paragraph{Continuous space-time models} Modeling dynamics in the continuum, whether in the spatial or temporal dimensions, with differential equations has become popular in deep learning \citep{lu2021learning,chamberlain2021grand,kovachki2023neural}. This approach is particularly convenient to operate in scenarios involving irregularly sampled data~\citep{shukla2020survey}, where both training and inference can be performed w.r.t\ arbitrary points in time and space \citep{chen2018neural,kidger2020neural}. Indeed, continuous space-time models find applications in 
multi-scale analysis and simulation of physical systems \citep{raissi2019physicsinformed}. 
\citet{rubanova2019latent} pioneered research in this direction showcasing the potential of such techniques in time-series applications. Graph-based approaches have been recently adopting an analogous approach to spatiotemporal data~\citep{huang2021coupled,fang2021spatialtemporala, choi2022graph, liu2023graphbased}, dynamic topologies~\citep{luo2023hope}, and graph learning procedures~\citep{jin2022multivariate}. Future works should address the design of sound and unifying frameworks for neural continuous space-time modeling; in particular, models that process space and time in an integrated fashion should be further explored.

\paragraph{Probabilistic forecasting} While we focused on point predictions, deep learning methods have been widely applied to probabilistic forecasting~\citep{ salinas2020deepar, wen2017multi, rangapuram2018deep, gasthaus2019probabilistic, debezenac2020normalizing}. One can in principle exploit~(most of) such methodologies to make \glspl{stgnn} output probabilistic predictions. As an example, quantile regression~\citep{koenker2001quantile} allows for obtaining uncertainty estimates by simply using an appropriate loss function and adding an output for each quantile of interest~\citep{wen2017multi}. In this regard, \citet{wu2021quantifying} carry out a study of several standard uncertainty estimation techniques in the context of spatiotemporal forecasting. However, while specialized probabilistic \gls{stgnn} architectures exist~(e.g,~\cite{pal2021rnn, chen2021graph}), as similarly recognized in~\citep{jin2023survey}, future works should further explore the use of relational inductive biases to obtain calibrated probabilistic forecasts and uncertainty estimates.

\paragraph{Open benchmarks and models} As mentioned in \autoref{sec:related}, graph deep learning for time series forecasting has been historically pioneered in the context of traffic forecasting and, as a consequence, most of the publicly available benchmarks come from this domain~(e.g., see~\cite{li2018diffusion, guo2021learning}). Within this context, \citet{liu2023largest} released one of the largest benchmarks available, consisting of a large collection of traffic speed measurements recorded by the California Department of Transportation\footnote{\href{https://pems.dot.ca.gov/}{https://pems.dot.ca.gov/}}. The \gls{cer} dataset~\citep{cer2016cer} has been widely used as a load forecasting benchmark~\cite{cini2022filling, marisca2022learning, cini2023taming}. \citet{cini2023scalable} introduced large-scale benchmarks for forecasting photovoltaic production and energy consumption, with the latter being based on \gls{cer}. Additional weather and photovoltaic forecasting datasets were introduced by~\citet{marisca2024graph} and~\citet{defelice2024graphbased}.  Other commonly used benchmarks come from air quality monitoring applications~\citep{zheng2015forecasting}. However, a large and heterogeneous benchmark for correlated time series forecasting is currently missing, as well as a shared benchmarking procedure and software infrastructure. Although standardized implementations of popular architectures are becoming available~\citep{Cini_Torch_Spatiotemporal_2022, liang2022basicts}, the field would benefit from shared evaluation platforms and benchmarks, e.g., following similar trends in~(temporal) graph learning~\citep{gravina2023deep,hu2020open, huang2023temporal}.

\section{Conclusions}\label{sec:conclusion}

We introduced a comprehensive methodological framework for time series forecasting with graph deep learning methods. We formalized the problem setup, characterizing settings common to several application domains. We then introduced different \acrlong{stgnn} architectures; we discussed their properties, advantages, and drawbacks with particular attention to the impact of global and local components. We then discussed ad-hoc techniques for model evaluation and performance assessment. We identified the challenges inherent to the field and the possible strategies to address them. Finally, we provided an outlook on future research directions. 

This paper offers a foundation for future research to build on, as well as a tutorial for practitioners. The result is a set of graph deep learning methods aimed at enriching modern time series forecasting practices and targeting practical, high-impact, real-world applications.

\bibliographystyle{ACM-Reference-Format}
\bibliography{bibliography}

\appendix

\appendix

\section*{Appendix}

\section{Software}\label{a:software}

PyTorch Geometric (PyG) \citep{fey2019fast} is the most widely used library for developing graph neural networks. As the name suggests, PyG is based on PyTorch~\citep{paske2019pytorch} and offers utilities to process temporal relational data as well. Specialized libraries that implement models from the temporal graph learning literature exist~\citep{rozemberczki2021pytorch}. The PyTorch ecosystem has several options for what concern deep learning for time series forecasting such as GluonTS~\citep{alexandrov2021gluonts}, PyTorch Forecasting\footnote{\href{https://github.com/jdb78/pytorch-forecasting}{https://github.com/jdb78/pytorch-forecasting}}, Neural Forecast~\citep{olivares2022library_neuralforecast} and BasicTS~\citep{liang2022basicts}. Considering the settings discussed in the paper, Torch Spatiotemporal~\citep{Cini_Torch_Spatiotemporal_2022} focuses on graph deep learning models for processing time series collections, offering several utilities to accelerate research and prototyping. 

\section{Experimental setting}

As mentioned in the paper, the setup of the computational experiments closely follows \citet{cini2023taming} for most of the considered scenarios. We report the pre-processing steps here for completeness. 
\subsection{Datasets}\label{a:datasets}

\begin{table*}[t]
\setlength{\aboverulesep}{0pt}
\setlength{\belowrulesep}{0pt}
\renewcommand{\arraystretch}{1.2}
\centering
\begin{small}
\begin{tabular}{l|c c c c c}
\toprule
 \sc Datasets & Type & Time steps & Nodes & Edges & Rate\\
\toprule
\gls{gpvar} & Undirected & 30,000 & 120 & 199 & N/A \\
\gls{lgpvar} & Undirected & 30,000 & 120 & 199 & N/A \\
\midrule
\gls{la} & Directed & 34,272 & 207 & 1515 & 5 minutes \\
\gls{bay} & Directed & 52,128 & 325 & 2369 & 5 minutes \\
\gls{cer} & Directed & 25,728 & 485 & 4365 & 30 minutes \\
\gls{air} & Undirected & 8,760 & 437 & 2699 & 1 hour \\
\bottomrule
\end{tabular}
\end{small}
\caption{Statistics of datasets used in the experiments.}\label{t:datasets}
\end{table*}

\autoref{t:datasets} and the following provide relevant additional information on each dataset.

\begin{figure}[t]
    \centering
    \includegraphics[width=.7\columnwidth]{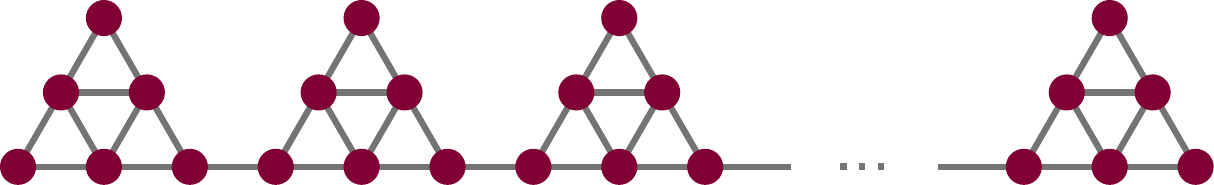}
    \caption{GPVAR community graph. We used a graph with $20$ communities resulting in a network with $120$ nodes.}
    \label{fig:gpvar-adj}
\end{figure}

\paragraph{GPVAR}  As already mentioned for the GPVAR datasets we follow the same setting reported in \citep{cini2023taming}. In particular, the parameters of the spatiotemporal process are set as
$$
\Theta = \left[\begin{smallmatrix}
  \hfill2.5 & \hfill-2.0 & \hfill-0.5\\
  \hfill1.0 & \hfill3.0 & \hfill0.0 \\ 
\end{smallmatrix}\right], \qquad \va,\vb \sim \mathcal{U}\left(-2, 2\right),
$$
$$
\eta \sim \mathcal{N}(\boldsymbol{0}, \text{diag}(\sigma^2)), \quad \sigma=0.4.
$$
The graph topology used to generate the data is the community graph shown in Fig.~\ref{fig:gpvar-adj}. In particular, we considered a network with $120$ nodes with $20$ communities and then added self-loops by setting the diagonal of the corresponding adjacency matrix to $1$. 

\paragraph{Benchmarks} We normalize the target variable to have zero mean and unit variance on the training set. The adjacency matrix for each dataset is obtained as discussed in \autoref{sec:experiments} by following previous works~\citep{cini2022filling,li2018diffusion,guo2021learning, marisca2022learning}. We use as exogenous variables sinusoidal functions encoding the time of the day and, for each dataset excluding \gls{engrad}, one-hot encodings of the day of the week. We split datasets into windows of $W$ time steps, and train the models to predict the next $H$ observations. Unless otherwise stated, the obtained windows are sequentially split into $70\%/10\%/20\%$ partitions for training, validation, and testing, respectively. In the following, we report detailed information for experiments on each dataset.
\begin{description}[leftmargin=1em, itemindent=1em]
    \item[\gls{la} \& \gls{bay}] Window and horizon length are set as $W=12$ and $H=12$. For \gls{la}, given a large number of missing values, we add as an additional exogenous variable the binary mask introduced in \autoref{sec:extension}.
    \item[\textbf{\gls{cer}}] Window and horizon length are set as $W=48$ and $H=6$.
    \item[\textbf{\gls{air}}] Window and horizon length are set as $W=24$ and $H=3$. We use the same data splits of previous works for training and evaluation~\citet{yi2016st, cini2022filling}.
    \item[\textbf{\gls{engrad}}] Window and horizon length are set as $W=24$ and $H=6$. We use the other weather variables from the dataset, along with sinusoidal encodings of the time of the year, as additional exogenous inputs. We normalize temperature values to have zero mean and unit variance. We do not compute loss and metrics on time steps with zero radiance and follow the protocol of previous work~\citet{marisca2024graph} to obtain training/validation/testing folds.
\end{description}

\paragraph{Hyperparameters} The reference \gls{tts} architectures are implemented by a single-layer GRU followed by $2$ message-passing layers. \glspl{gcrnn} have a single layer as well. 
For the benchmark datasets, the number of neurons in each layer is set to $64$ and the embedding size to $32$ for all the reference architectures and the \gls{rnn} baselines. We train with early stopping for a maximum of $200$ epochs with the Adam optimizer~\citep{kingma2014adam} and a learning rate of $0.003$ divided by four every $50$ epochs. In each training epoch, we randomly sample without replacement $300$ batches of size $64$ from the training set. We reduce both batch size and size of the hidden representations to $32$ when the model exceeds the available GPU memory capacity~(approximately $24$ GB). 
For GPVAR, we use $16$ and $8$ as hidden and embedding sizes, respectively, and $128$ as the batch size. We set  $0.01$ as the initial learning rate, which is halved every $50$ epochs.

\end{document}